\documentclass{article}

\usepackage{arxiv}

\usepackage[utf8]{inputenc} 
\usepackage[T1]{fontenc}    
\usepackage{hyperref}       
\usepackage{url}            
\usepackage{booktabs}       
\usepackage{amsfonts}       
\usepackage{nicefrac}       
\usepackage{microtype}      
\usepackage{lipsum}
\usepackage{authblk}

\usepackage{balance}
\usepackage[numbers]{natbib}
\usepackage{float} 
\usepackage{times}
\usepackage{epsfig}
\usepackage{amsmath, amsfonts, amssymb}
\usepackage{bm}
\usepackage{algorithm}
\usepackage{algorithmicx}
\usepackage{comment}
\usepackage{enumitem}
\usepackage{fullpage}
\usepackage{tikz}
\usepackage{pifont}
\usepackage{amsthm}
\usepackage{subcaption}
\usepackage{mathtools}
\usepackage{multirow}
\usepackage[noend]{algpseudocode}
\usepackage{booktabs}
\usepackage[export]{adjustbox}
\usepackage{url}

\usepackage{xcolor}
\hypersetup{
	colorlinks=true,
	linkcolor=blue,
	citecolor=blue,
	urlcolor=blue
}

\usepackage[textsize=scriptsize,backgroundcolor=green]{todonotes}

\newtheorem*{remark}{Remark}

\newcommand{\cmark}{{\color{blue}\ding{51}}}%
\newcommand{\xmark}{{\color{red}\ding{55}}}

\newcommand\numberthis{\addtocounter{equation}{1}\tag{\theequation}}

\usepackage{xspace}
\newcommand*{\eg}{\textit{e.g.}\@\xspace}
\newcommand*{\ie}{\textit{i.e.}\@\xspace}
\newcommand*{\vs}{\textit{vs.}\@\xspace}
\newcommand*{\wrt}{\textit{w.r.t.}\@\xspace}
\makeatletter
\newcommand*{\etc}{%
	\@ifnextchar{.}%
	{\textit{etc}}%
	{\textit{etc.}\@\xspace}%
}
\makeatother
\algtext*{EndWhile}
\algtext*{EndFor}
\algtext*{EndIf}
\algdef{SE}[DOWHILE]{Do}{doWhile}{\algorithmicdo}[1]{\algorithmicwhile\ #1}%

\makeatletter
\def\BState{\State\hskip-\ALG@thistlm}
\makeatother

\begin{document}

\title{Scalable Unidirectional Pareto Optimality for Multi-Task Learning with Constraints}

\author[1,4]{\textbf{Soumyajit Gupta} \textsuperscript{\dag}}
\author[5]{\textbf{Gurpreet Singh} \textsuperscript{\dag}}
\author[2,4]{\textbf{Raghu Bollapragada}}
\author[3,4]{\textbf{Matthew Lease}}
\affil[1]{Department of Computer Science}
\affil[2]{Operations Research and Industrial Engineering}
\affil[3]{School of Information}
\affil[4]{The University of Texas at Austin}
\affil[5]{XtractorAI}
\affil[ ]{\texttt{\{smjtgupta,raghu.bollapragada,ml\}@utexas.edu,gurpreet@xtractorai.com}}
\renewcommand\Authands{ and }

\maketitle

{\let\thefootnote\relax\footnote{{\dag contributed equally to this work.}}}

\begin{abstract}
Multi-objective optimization (MOO) problems require balancing competing objectives, often under constraints. The {\em Pareto optimal} solution set defines all possible optimal trade-offs over such objectives. In this work, we present a novel method for  {\em Pareto-front learning}: inducing the full Pareto manifold at train-time so users can pick any desired optimal trade-off point at run-time. Our key insight is to exploit Fritz-John Conditions for a novel guided {\em double gradient descent} strategy. Evaluation on synthetic benchmark problems allows us to vary MOO problem difficulty in controlled fashion and measure accuracy \vs known analytic solutions. We further test scalability and generalization in learning optimal neural model parameterizations for Multi-Task Learning (MTL) on image classification. Results show consistent improvement in  accuracy and efficiency over prior MTL methods as well as techniques from operations research.
\end{abstract}






\section{Introduction}

Multi-Objective Optimization (MOO) problems require balancing multiple objectives, often competing with one another under further constraints \citep{van1994saddle,ehrgott2005saddle}. A {\em Pareto optimal} solution \citep{pareto1906manuale}  defines the set of all saddle points \citep{ehrgott2005saddle} such that no objective can be further improved without penalizing at least one other objective. 

As operational systems today increasingly seek to balance competing objectives, research on Pareto optimal learning has quickly grown across tasks such as fair classification \citep{balashankar2019fair,martinez2020minimax}, diversified ranking \citep{liu2019skyrec,sacharidis2019top}, and recommendation \citep{xiao2017fairness,azadjalal2017trust}. 
Many practical classification and recommendation problems have been shown to be non-convex \citep{hsieh2015pu}. A general Pareto solver should thus support optimization for both non-convex objectives and constraints. 

Because MOO problems typically lack a single global optimum, one must choose among optimal solutions by selecting a trade-off over competing objectives. Ideally this choice could be deferred to run-time, so that each user could choose whichever trade-off they prefer. Unfortunately, prior Pareto solvers have typically required training a separate model to find the Pareto solution point for each desired trade-off.


To address this, recent work has proposed {\em Pareto front learning} (PFL): inducing the full Pareto manifold in training so that users can quickly select any desired optimal trade-off point at run-time \citep{navon2021learning,lin2021controllable,singh2021hybrid}. These works learn a neural model manifold to map any desired trade-off over objectives to a corresponding Pareto point. As with other supervised learning, inducing 
an accurate prediction model requires high quality training data, \ie, Pareto points used for training should be accurate.


In this work, we devise a efficient Pareto search procedure for \citet{singh2021hybrid}'s HNPF model, so that we may benefit from its correctness guarantees in identifying true Pareto points for PFL training. While HNPF supports non-convex MOO with constraints and bounded error, it suffers from a lack of scalability with increasing variable space. Our innovation is a novel,  
guided {\em double gradient descent} strategy, updating the candidate point set in the outer descent loop and the manifold estimators in the inner descent loop. 

Our evaluation spans both synthetic benchmarks and multi-task learning (MTL) problems. Benchmark problems allow us to conduct controlled experiments varying MOO problem complexity (\eg the presence of constraints and/or convexity in variable or function domains). Analytic solutions to benchmark problems allow us to measure the true accuracy of model predictions, something which is often difficult or impossible on real-world problems. Additional evaluation on a set of MTL problems in image classification enable us to further test scalability and generalization in learning Pareto optimal  weights for high dimensional neural models.

Results across synthetic benchmarks and MTL problems show clear, consistent advantages of SUHNPF in terms of capability (handling non-convexity and constraints), denser coverage and higher accuracy in recovering the true Pareto front, and greater efficiency (time and space). Beyond empirical findings, our conceptual framing and review of prior work also serves to further bridge complementary lines of MTL and operations research work. For reproducibility, we will share our sourcecode and data upon publication.


\section{Definitions}

We adopt Pareto definitions from \cite{marler2004survey}. A general MOO problem can formulated as follows:
\begin{align*}
	&\underset{}{\textrm{optimize}} \quad F(x) = (f_1(x),\ldots,f_k(x)) \numberthis \label{eq:multi}\\
	&\text{s.t.} \quad x \in S = \{ x \in \mathbb{R}^n | G(x)=(g_1(x),\ldots,g_m(x)) \leq 0 \}
\end{align*}

with $n$ variables $(x_1,\ldots,x_n)$, $k$ objectives $(f_1,\ldots,f_k)$, and $m$ constraints $(g_1,\ldots,g_m)$. Here, $S$ is the feasible set, \ie the set of input values $x$ that satisfy the constraints $G(x)$. For a MOO problem optimizing $F(x)$ subject to $G(x)$, the solution is usually a manifold as opposed to a single global optimum, therefore one must find the set of all points that satisfy the chosen definition for an optimum. 

\textbf{Strong Pareto Optimal:} A point $\tilde{x}^* \in S$ is {\em strong} Pareto optimal if no point in the feasible set exists that improves an objective without detriment to at least one other objective.
\begin{align*}
	&\nexists x_j: f_p(x_j) \leq f_p(x^*), \quad \textrm{for} \quad p=1,2,\ldots,k\\
	&\exists l: f_l(x_j) < f_l(x^*) \numberthis \label{eq:pareto}
\end{align*}

\textbf{Weak Pareto Optimal:} A point $\tilde{x}^* \in S$ is {\em weak} Pareto optimal if no other point exists in the feasible set that improves all of the objectives simultaneously. This is different from strong Pareto, where points might exist that improve at least one objective without detriment to another.
\begin{align}
	\nexists x_j: f_p(x_j) < f_p(\tilde{x}^*), \quad \textrm{for} \quad p=1,2,\ldots,k
\end{align}



\section{Related Work} \label{sec:related}


\textbf{Linear Scalarization (LS)}. A variety of work has adopted LS to find Pareto points \citep{xiao2017fairness,lin2019pareto,milojkovic2019multi}. For example, the Weighted Sum Method (WSM) \cite{cohon2004multiobjective} is a LS approach to convert an MOO into a SOO using a convex combination of objective functions and constraints. However, because Karush-Kuhn-Tucker (KKT) conditions are known to hold true only for convex cases \citep{boyd2004convex}, LS solutions are guaranteed to be Pareto optimal only under fully convex setting of objectives and constraints, as shown in \cite{gobbi2015analytical}.

\textbf{Operations Research (OR)}. A variety of OR methods support MOO problems with non-convex objectives and constraints, guaranteeing correctness within a user-specified error tolerance. Correctness has also been further verified by evaluation on synthetic MOO benchmark problems with known, analytic solutions. However, a key limitation of these methods is lack of scalability: they suffer from significant computational and run-time limitations as the variable dimension increases. Hence, they cannot be applied to optimizing neural model parameters for MOO problems.

\begin{table}[bht]
	\centering
	\caption{\small SUHNPF \vs existing Operations Research (OR) and Multi-Task Learning (MTL) methods. OR methods account for both objectives and constraints, produce Pareto points only, and are known to find true Pareto points for non-convex MOO problems. 
		However, these methods do not scale to high-dimensional neural MOO problems. In contrast, MTL methods scale well but typically do not support constraints and can struggle with non-convexity.} 
	\begin{tabular}{ll|ccc}
		\toprule
		\bf Type & \bf Method & \bf Finds Only  & \bf Handles & \bf  Scalable \\ 
		& & \bf \!\!Pareto points\!\! & \bf  \!\!Constraints\!\! & \bf \!\!Neural MOO\!\!\!\! \\ \midrule
		\multirow{4}{40pt}{Operations Research (OR)}        
		& NBI [\citeyear{das1998normal}] & \cmark & \cmark & \xmark \\
		& mCHIM [\citeyear{ghane2015new}] & \cmark & \cmark & \xmark \\
		& PK [\citeyear{pirouz2016computational}] & \cmark & \cmark & \xmark \\
		& HNPF [\citeyear{singh2021hybrid}] & \cmark & \cmark & \xmark \\ \midrule
		\multirow{4}{40pt}{Multi-Task Learning (MTL)} 
		& MOOMTL [\citeyear{sener2018multi}]\!\!\! & \xmark & \xmark & \cmark \\
		& PMTL [\citeyear{lin2019pareto}] & \xmark & \xmark & \cmark \\
		& EPO [\citeyear{mahapatra2020multi}] & \xmark & \xmark & \cmark \\
		& EPSE [\citeyear{ma2020efficient}] & \xmark & \xmark & \cmark \\
		& PHN [\citeyear{navon2021learning}] & \xmark & \xmark & \cmark \\ \midrule
		Ours & \bf SUHNPF & \cmark & \cmark & \cmark \\ \bottomrule        
\end{tabular}
\label{tab:comp}
\end{table}

Examples include enhanced scalarization approaches such as NBI \citep{das1998normal}, mCHIM \citep{ghane2015new}, and PK \citep{pirouz2016computational}.
NBI produces an evenly distributed set of Pareto points given an evenly distributed set of weights,  
using the concept of Convex Hull of Individual Minima (CHIM) to break down the boundary/hull into evenly spaced segments before tracing the {\em weak} Pareto points. mCHIM improves upon NBI via a quasi-normal procedure to update the aforementioned CHIM set iteratively, to obtain a strong Pareto set. PK uses a local $\epsilon$-scalarization based strategy that searches for the Pareto front using controllable step-lengths in a restricted search region, thereby accounting for non-convexity. 

\textbf{Multi-Task Learning (MTL)}. Recent MTL works 
have devised Pareto solvers for estimating high-dimensional neural models. 
%
%
MOOMTL \citep{sener2018multi} effectively scales via a multi-gradient descent approach, but does not guarantee an even spread of solution points found along the Pareto front. PMTL \citep{lin2019pareto} addresses this spread issue by diving the functional domain into equal spaced cones, but this increases computational complexity as the number of cones increases. EPO \citep{mahapatra2020multi} extends preference rays along specified weights to find Pareto points evenly spread in the vicinity of the rays. EPSE \citep{ma2020efficient} uses a combination Hessian of the functions and Krylov subspace to find Pareto solutions. 

MTL methods rely upon KKT 
conditions 
to check for optimality, which assumes convexity (see earlier LS discussion). While methods seek an even distribution of Pareto points by dividing the functional space into evenly spaced cones or preference rays, 
our results on a non-convex benchmark problem clearly show an uneven point spread (\textbf{Section~\ref{sec:case1}}). 
%
Moreover, most MTL methods 
are {\em point-based solvers}, meaning they must be run $P$ times to find $P$ points. 
This is too expensive to adjust trade-off preferences at run-time. 

{\bf Pareto front learning}. PFL methods \citep{navon2021learning,lin2021controllable,singh2021hybrid} induce  the  full  Pareto  manifold  at  train-time so that users can quickly select any desired optimal trade-off point at run-time. For example, a manifold model trained on $P$ Pareto points might then quickly produce any number of additional Pareto points via interpolation. Of course, quality training data quality is necessary to learn an accurate, supervised prediction model. The method and resulting accuracy of the Pareto points used for model training is thus crucial to prediction accuracy.


\citet{navon2021learning}'s PHN considers two way to acquire Pareto training points: LS and EPO [\citeyear{mahapatra2020multi}]. \citet{lin2021controllable} use their PMTL [\citeyear{lin2019pareto}] method to identify Pareto points for training. \citet{singh2021hybrid}'s HNPF uses the Fritz-John conditions (FJC) \citep{marucsciac1982fritz} to identify Pareto points.

Like other OR methods, HNPF provides a theoretical guarantee of Pareto front accuracy within a user-specified error tolerance.  In evaluation on canonical OR benchmark problems, HNPF was shown to recover known Pareto fronts across various non-convex MOO problems while also being more efficient in finding Pareto points than NBI [\citeyear{das1998normal}], mCHIM [\citeyear{ghane2015new}], and PK [\citeyear{pirouz2016computational}]). However, like other OR methods, HNPF cannot scale to learn optimal high-dimensional neural model weights for MOO problems. 


\citet{ha2016hypernetworks}'s hypernetworks proposed training one neural model to generate effective weights for a second, target model. \citet{navon2021learning} and \citet{lin2021controllable} apply this approach to learn a manifold mapping MOO solutions to different target model weights, enabling the target model to achieve the desired Pareto trade-off for the MOO problem. 
However, HNPF cannot be similarly applied to MTL problems due to its lack of scalability. 

\section{Preliminaries}

{\bf Fritz John Conditions (FJC)}. Let the objective and constraint function in Eq. \eqref{eq:multi} be differentiable once at a decision vector $x^* \in \mathcal{S}$. The Fritz-John \citep{levi2006application} necessary conditions for $x^*$ to be {\em weak} Pareto optimal is that vectors must exists for $0 \leq \lambda \in \mathbb{R}^k$, $0 \leq \mu \in \mathbb{R}^m$ and $(\lambda, \mu) \neq (0,0)$ (not identically zero) \textit{s.t.} the following holds:
\begin{align*}
\sum_{i=1}^k \lambda_i \nabla f_i(x^*) + \sum_{j=1}^m \mu_j \nabla g_j(x^*) = 0 \numberthis \label{eq:fjcond} \\
\mu_jg_j(x^*) = 0, \forall j=1,\ldots,m
\end{align*}

\citet{gobbi2015analytical} present an $L$ matrix form of FJC:
\begin{align*}
&L = \begin{bmatrix}
	\nabla F & \nabla G \\
	\mathbf{0} & G
\end{bmatrix} \quad [(n+m) \times (k+m)] \label{eq:fjmat} \numberthis \\
&\nabla F_{n \times k} = [\nabla f_1, \ldots, \nabla f_k]\\
&\nabla G_{n \times m} = [\nabla g_1, \ldots, \nabla g_m]\\
&G_{m \times m}=diag(g_1,\ldots,g_m) 
\end{align*}

comprising the gradients of the functions and constraints. The matrix equivalent of FJC for $x^*$ to be Pareto optimal is to show the existence of $\delta = (\lambda, \mu) \in \mathbb{R}^{k+m}$ (\ie $\mathbf{\delta}$ not identically zero) in Eq. \eqref{eq:fjcond} such that:
\begin{align}
L \cdot \delta = 0 \quad \text{s.t.} \quad L=L(x^*),\mathbf{\delta} \geq 0, \mathbf{\delta} \neq 0 \label{eq:fjmatrix}
\end{align}
Therefore the non-trivial solution for Eq. \eqref{eq:fjmatrix} is:
\begin{align}
det(L^TL)=0 \label{eq:paropt}
\end{align}

\begin{remark}
\small If $f_i$s and $g_j$s are continuous and differentiable once, then the set of weak Pareto optimal points are $x^*=\{x|det(L(x)^TL(x))=0\}$, $\delta \geq 0$ for a non-square matrix $L(x)$, and is equivalent to $x^*=\{x|det(L(x))=0\}$, $\delta \geq 0$, for a square matrix $L(x)$. \textup{See an illustration in \textbf{Appendix \ref{app:fjc}}} for the unconstrained setting.
\end{remark}

\textbf{Hybrid Neural Pareto Front (HNPF)}. 
Like other Pareto front learning (PFL) methods, HNPF \citep{singh2021hybrid} learns a neural Pareto manifold from training data. With HNPF, Pareto points for use as training data data are acquired via Fritz-John conditions. In particular, once a given a data point from the input variable domain is mapped to the output function domain (via objective functions), FJC
are tested to determine Pareto optimality. 



HNPF's neural network first identifies {\em weak Pareto} points via feed-forward layers to smoothly approximate the \textit{weak} Pareto optimal solution manifold $M(X^*)$ as $\tilde{M}(\tilde{X},\Phi)$. 
The last layer of the network has two neurons with \textit{softmax} activation for binary classification of Pareto \vs non-Pareto points, distinguishing sub-optimal points from the {\em weak} Pareto points. 
The network loss is representation driven, since the Fritz John discriminator (Eq. \eqref{eq:paropt}), described by the objective functions and constraints, explicitly classifies each input data point $X_i$ as being {\em weak} Pareto or not.  
%
%
%
%
After identifying weak Pareto points, HNPF uses an efficient Pareto filter to find the subset of \textit{non-dominated} points. 

HNPF's scalability bottleneck lies in how it samples variable domain points to test for Pareto optimality in model training. If there are any direct constraints on variable values, this naturally restricts the feasible domain for sampling. However, lacking any prior distribution on where to find Pareto optima, HNPF performs uniform random sampling in the variable domain to ensure broad coverage for locating optima. For small benchmark problems with known variable domains, this suffices. However, it is infeasible to apply this to find optimal model parameters for a neural MOO model.

\section{Scalable Unidirectional HNPF}

To address HNPF's scalability bottleneck, we introduce SUHNPF, a scalable variant of HNPF for finding weak Pareto points with an arbitrary density and distribution of initial data points. This is achieved via a scalable unidirectional FJC-guided double-gradient descent algorithm that encompasses HNPF's neural manifold estimator. Given continuous differentiable loss functions, SUNHPF's guided double gradient descent strategy efficiently searches the variable domain to find Pareto optimal points in the function domain. This enables SUHNPF to learn an $\epsilon$-bounded approximation $\tilde{M}(\Theta^*)$ to the weak Pareto optimal manifold. 


\subsection{FJC-Guided Double Gradient Descent}

Constructing a classification manifold of Pareto \vs non-Pareto points requires a set of feasible points to represent both classes. Since the Pareto manifold is unknown \textit{a priori}, feasible points are drawn from a random distribution (lacking an informed prior) to initialize both classes. 
We then refine the points in the Pareto class $\mathcal{P}1$ while holding the non-Pareto points $\mathcal{P}0$ constant.



We assume an equal-sized sample set of $P$ points for each class, which helps to address class imbalance for harsh cases. For benchmark problems where the feasible set over the variable domain is known, we randomly sample points over this feasible domain to initialize $\mathcal{P}1$ and $\mathcal{P}0$. Given these input points $x$, held constant for $\mathcal{P}0$ and used as initial seed values for $\mathcal{P}1$, \textbf{Alg. \ref{alg:fjc}} specifies our FJC-guided double-gradient descent algorithm. The algorithm iteratively 
updates $\mathcal{P}1$ towards the Pareto manifold via FJC-guided descent. The training dataset $D$ is the union of $\mathcal{P}0 \cup \mathcal{P}1$. The algorithm iterates over Steps 5-9 until the error ($err$) converges to the user-specified error tolerance ($\epsilon_{outer}$).
\begin{equation}
err = \sum_{p \in \mathcal{P}1} \left (det(L^TL) \right )^2 \label{eq:detloss}
\end{equation}


\begin{algorithm}[htb]
\small
\caption{FJC-guided descent of variable domain}
\begin{algorithmic}[1]
	\BState \textbf{Input}: Data $D = \mathcal{P}0 \cup \mathcal{P}1$ \Comment{Training Data}
	\BState \textbf{Input}: Functions $F$ and Constraints $G$
	\BState \textbf{Input}: Error tolerance $\epsilon_{outer}$, $\epsilon_{inner}$
	\While {$err > \epsilon_{outer}$} \Comment{Run until convergence}
	\State Train network using $D$ as data for $e$ epochs
	\State Compute current error $err$ 
	\State Compute $\nabla_{p} det = \frac{\partial det(L^TL)}{\partial p}$, $\forall p \in \mathcal{P}1$
	\State $\mathcal{P}1 \leftarrow \mathcal{P}1 - \eta \nabla det$ \Comment{Update points in $\mathcal{P}1$}
	\State $D = \mathcal{P}0 \cup \mathcal{P}1$ \Comment{Update Training Data}
	\EndWhile
	\BState \textbf{Output}: Weak Pareto manifold $\tilde{M}$ 
\end{algorithmic} \label{alg:fjc}
\end{algorithm}

Eq. \ref{eq:detloss} in Alg.~\ref{alg:fjc} ensures that all of the points in the Pareto set ($p \in \mathcal{P}1$) are optimal once we converge to the desired error tolerance $\epsilon$. Hence, Step 7 computes gradients of the $det(L^TL)$ matrix \wrt the variables at points $p \in \mathcal{P}1$ and creates an approximation of the $\nabla det$ matrix. 
The training data $D$ is then updated with the new values of $\mathcal{P}1$. 
The output is an approximation of the true weak Pareto manifold $M$ as $\tilde{M}$ on the discrete dataset $D \subset X$. Note that in Step 8, we do not allow the point set $\mathcal{P}1$ to leave the feasible set $\mathcal{S}$ \ie if the step crosses the boundary of the feasible set, then we update the point to be the point on the boundary.

Alg. \ref{alg:fjc} includes two separate gradient descent steps. The outer descent loop (Step 4-9) updates the candidate point set $\mathcal{P}1$ using the error measurement of $err$ through a squared loss in Eq. \ref{eq:detloss}. The inner descent (Step 5) updates the parameters ($\Phi$) of the neural net to closely approximate the Pareto manifold $M(X)$ as $\tilde{M}(X,\Phi)$. This is done using the Binary Cross Entropy Loss on ($det(L(X)^TL(X)),\tilde{M}(X)$), and reaches convergence only when $BCE \leq \epsilon_{inner}$. The \textit{unidirectional} property of this double-gradient update lets the outer loop influence the inner loop but not vice-versa.

{\bf Space Complexity Analysis}. Alg.~\ref{alg:fjc} maintains $D$ of size ${P \times n}$, with $L^TL$ and $\nabla det$ matrices of sizes $(k+m)^2$ and $n(k+m)$, respectively. Thus $O(n(k+m+P)+(k+m)^2)$ total memory is used, where $n$ is the dimension of the variable space, and the scale of $k,m,P$ varies \wrt the problem. Trade-off $\alpha$'s are computed by solving a linear system as post-processing. SUHNPF achieves better memory and run-time efficiency since it does not rely upon solving primal and dual problems used in MTL methods (\textbf{Appendix \ref{app:space}}).


\section{Benchmarking} \label{sec:benchmark}

{\bf Motivation.} Lack of analytical solutions to real MOO problems makes it difficult to measure the true accuracy of any Pareto solver. Consequently, we follow the OR literature in advocating that the correctness of any proposed Pareto solver should first be tested on constructed benchmark problems with known analytic solutions. This is also consistent with broader ML community practice of first evaluating proposed methods across a range of simulated, controlled conditions to verify correctness, often yielding valuable insights into model behavior prior to evaluation on real data.


We consider three such benchmark problems (Cases I-III). These problems are non-convex in either the functional or variable domain, or due to constraints (\textbf{Table \ref{tab:cases}}). Note that {\em whether or not the Pareto front itself is non-convex is not always the best indicator of benchmark difficulty}. For example, even though both objectives are non-convex in Case II, the Pareto front is still convex. As we shall see, PHN \citep{navon2021learning} fails on Case II despite performing well on two benchmark problems in their own study having a non-convex front. In general, non-convexity can greatly challenge MTL approaches relying on KKT conditions in testing solutions for optimality (see \textbf{Appendix \ref{app:discussion}}).

\begin{table}[bht]
\centering
\caption{\small Characterization of benchmark cases, including convexity (C) \vs non-convexity (NC) in variable and function domains.}
\begin{tabular}{r|rccc|ccc}
	\toprule
	\bf \!\!Case\!\! & \!\!{\bf Dim}\!\! & \begin{tabular}[c]{@{}c@{}}\bf Variable\\ \bf Domain\end{tabular} & \begin{tabular}[c]{@{}c@{}}\bf Function\\\bf Domain\end{tabular} & \begin{tabular}[c]{@{}c@{}}\bf Includes\\ \bf \!\!Constraints\!\!\end{tabular} & \begin{tabular}[c]{@{}c@{}}\bf OR\\\bf Methods\end{tabular} & \begin{tabular}[c]{@{}c@{}}\bf MTL\\\bf Methods\end{tabular} & \bf SUHNPF \\ \midrule
	I & 2 & Linear & NC & No & Sparse, Slow & \!\!\!Sparse, Fast\!\!\! & Dense, Fast \\
	II & 30 & NC & C & No & Sparse, Slow & Fail & Dense, Fast \\
	III & 2 & NC & NC & Yes & Sparse, Slow & Fail & Dense, Fast \\ \bottomrule        
\end{tabular}
\label{tab:cases}
\end{table}

{\bf Experimental Setup}. For each Case I-III, each method is tasked with finding $P=50$ Pareto points. OR methods search until any $P$ Pareto points are found. MTL methods divide the functional search quadrant into cones/rays, seeking one Pareto point per split. Manifold-based methods (PHN, HNPF, and SUHNPF) search for $P$ Pareto points in order to learn the manifold. Ideally, each method should identify an even spread (\ie broad coverage) of points across the true Pareto front (shown in grey in each figure) in order to faithfully approximate it. We report the number of iterative steps (\ie evaluations) taken by each solver to find the points.


SUHNPF starts with $P$ random candidates that are progressively refined via its guided, double gradient descent strategy. Following HNPF \citep{singh2021hybrid}, we adopt the same error tolerance $10^{-4}$ for both $\epsilon_{outer}$ and $\epsilon_{inner}$. Any point $x$ that satisfies $|det(L(x)^TL(x))| \leq \epsilon_{inner}$ is thus classified as being Pareto (exact zero is often impossible given finite machine precision). Sourcecode for LS, MOOMTL, PMTL and EPO solvers are taken from EPO's repository, while EPSE and PHN's sourcecode are used for them, respectively (see \textbf{Appendix \ref{app:setup}}). Based on \cite{navon2021learning}'s findings, we evaluate the more accurate PHN variant, PHN-EPO, which we refer to simply as PHN. 

Due to key differences between OR \vs MTL methods, results for each group are presented separately. First, OR methods not only support the full range of non-convex conditions across Cases I-III, but provide error tolerance parameters to guarantee correctness (and our experiments confirm this). Consequently, we report only the efficiency of OR  methods in Table \ref{tab:evalsOR}. In contrast, MTL methods produced variable accuracy on Case I and failed entirely on Cases II and III (as shall be discussed). Consequently, Table \ref{tab:evalsMTL} reports accuracy and efficiency of MTL methods for Case I only. 

\textbf{Appendix~\ref{app:setup}} discusses experimental setup and \textbf{Appendix~\ref{app:loss}} has loss profiles.

\subsection{Case I: \cite{ghane2015new}} \label{sec:case1}

\resizebox{\linewidth}{!}{
\begin{minipage}{\linewidth}
\begin{align*}
	&f_1(x_1,x_2) = x_1 , \, f_2(x_1,x_2) = 1 + x_2^2 - x_1 - 0.1sin 3 \pi x_1\\
	&\text{s.t.} \quad g_1: 0 \leq x_1 \leq 1, g_2: -2 \leq x_2 \leq 2
\end{align*}
\end{minipage}
}

\begin{figure}[ht]
\centering
\begin{subfigure}{0.3\linewidth}
\centering
\includegraphics[width=\linewidth]{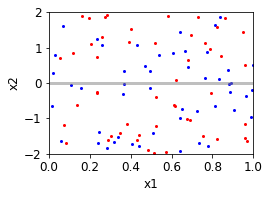}
\caption{Iteration 0 (Start)}
\end{subfigure}
\begin{subfigure}{0.3\linewidth}
\centering
\includegraphics[width=\linewidth]{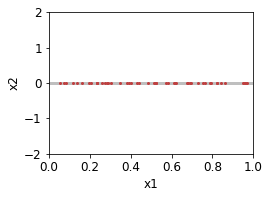}
\caption{Iteration 5 (Converged)}
\end{subfigure}
\caption{\small Case I: Variable domain. The gray line show the true analytic solution ($0 \leq x_1 \leq 1$). SUHNPF Pareto candidates $\mathcal{P}1$ (red dots) converge in 5 iterations. Non-Pareto candidates $\mathcal{P}0$ (blue dots) are held constant throughout the iterative sequence.}
\label{fig:illus-var}
\end{figure}

\begin{figure}[ht]
\centering
\begin{subfigure}{0.3\linewidth}
\centering
\includegraphics[width=\linewidth]{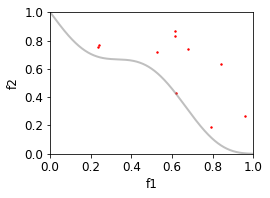}
\caption{Iteration 0 (Start)}
\end{subfigure}
\begin{subfigure}{0.3\linewidth}
\centering
\includegraphics[width=\linewidth]{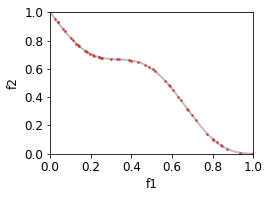}
\caption{Iteration 5 (Converged)}
\end{subfigure}
\caption{\small Case I: Functional domain corresponding to Figure \ref{fig:illus-var}. SUHNPF Pareto candidates $\mathcal{P}1$ (red dots) converge in 5 iterations.}
\label{fig:illus-func}
\end{figure}

The analytical Pareto solution to this joint minimization problem is $M: 0 \leq x_1 \leq 1,x_2=0$. In \textbf{Fig. \ref{fig:illus-var}} we observe SUHNPF's randomly generated point set $\mathcal{P}1$ (red dots) converges towards the true manifold $M$ as a discrete approximation $\tilde{M}$. Point set $\mathcal{P}0$ (blue dots) is held constant and serves as representatives for the (background) non-Pareto class. Iteration 5 is the last because the error falls below the user-specified $\epsilon$. The final cardinality of the weak Pareto set $|\mathcal{P}1| = P$ and any $\mathcal{P}0$ point that happens to fall within the $\epsilon_{outer}$ threshold. Hence Alg. \ref{alg:fjc} ensures 100\% Pareto point density in $\mathcal{P}1$, a vast improvement from HNPF \citep{singh2021hybrid}, where only $\approx$ 2\% density was achieved. 
\textbf{Fig. \ref{fig:illus-func}} shows functional domain convergence. 
SUHNPF achieves an even spread of points in the non-convex portion of the front.

\begin{figure}[ht]
\centering
\begin{subfigure}{0.3\linewidth}
\centering
\includegraphics[width=\linewidth]{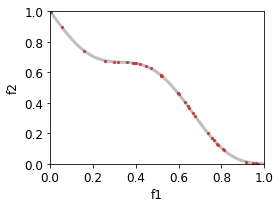}
\caption{Linear Scalarization (LS)}
\end{subfigure}
\begin{subfigure}{0.3\linewidth}
\centering
\includegraphics[width=\linewidth]{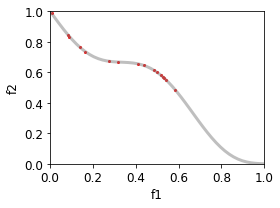}
\caption{MOOMTL}
\end{subfigure}
\begin{subfigure}{0.3\linewidth}
\centering
\includegraphics[width=\linewidth]{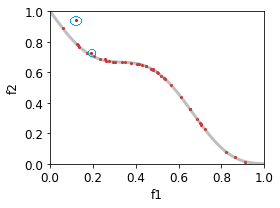}
\caption{PMTL}
\end{subfigure}
\begin{subfigure}{0.3\linewidth}
\centering
\includegraphics[width=\linewidth]{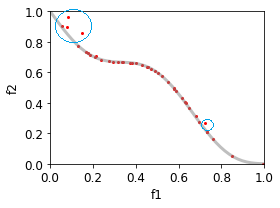}
\caption{EPO}
\end{subfigure}
\begin{subfigure}{0.3\linewidth}
\centering
\includegraphics[width=\linewidth]{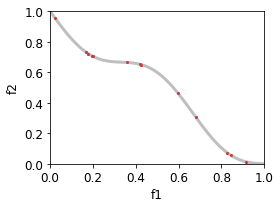}
\caption{EPSE}
\end{subfigure}
\begin{subfigure}{0.3\linewidth}
\centering
\includegraphics[width=\linewidth]{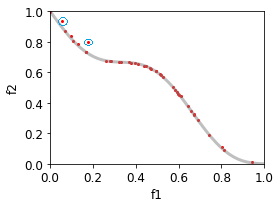}      
\caption{PHN}
\end{subfigure}
\caption{\small Case I: function domain for LS and MTL methods. No method produces all $50$ of the requested Pareto points. PMTL, EPO and PHN also find non-Pareto points (circled in blue). Methods vary greatly in their coverage of points spanning the true front.}
\label{fig:case1-mtl}
\end{figure}

\textbf{Fig. \ref{fig:case1-mtl}} presents results for Linear Scalarization (LS) and several MTL methods: MOOMTL, PMTL, EPO, EPSE, and PHN. LS successfully produces a number of points in the non-convex portions of the front, despite prior studies 
often asserting that LS cannot handle any non-convexity. Refer to \textbf{Appendix \ref{app:discussion}} for analysis and justification. 

To check for optimality, MTL methods rely upon KKT conditions that implicitly assume convexity (see Section \ref{sec:related}). The non-convex nature of $f_2$ is thus challenging for these KKT-based methods. For example, some methods seek an even distribution of Pareto points by breaking up the functional space into evenly spaced cones or preference rays for trade-off values $\alpha$. However, the uneven point spread seen on this non-convex benchmark illustrates limitations of the cone-based approach in handling non-convexity. We also clearly see non-Pareto points produced by some methods. 

\subsection{Case II: \cite{zhang2008multiobjective}}

\resizebox{\linewidth}{!}{
\begin{minipage}{\linewidth}
\begin{align*}
&f_1(x) = x_1 + \frac{2}{|J_1|}\sum_{j \in J_1}y_j^2 \hspace{1em},\hspace{1em} f_2(x) = 1 - \sqrt{x_1} + \frac{2}{|J_2|}\sum_{j \in J_2}y_j^2 \\
&\text{s.t.} \quad g_1,\ldots,g_{30}: 0 \leq x_1 \leq 1, -1 \leq x_j \leq 1, j=2,\ldots,m\\
& J_1=\{j|j \, \textrm{is odd},2 \leq j \leq m\},J_2=\{j|j \, \textrm{is even},2 \leq j \leq m\}\\
& y_j = \left\{\begin{matrix}
x_j - [0.3x_1^2 \cos(24\pi x_1 + \frac{4j\pi}{m}) + 0.6x_1] cos(6\pi x_1 + \frac{j\pi}{m}) \quad j \in J_1   \\ 
x_j - [0.3x_1^2 \cos(24\pi x_1 + \frac{4j\pi}{m}) + 0.6x_1] cos(6\pi x_1 + \frac{j\pi}{m}) \quad j \in J_2
\end{matrix}\right.
\end{align*}
\end{minipage}
}

\begin{figure}[ht]
\centering
\begin{subfigure}{0.3\linewidth}
\centering
\includegraphics[width=\linewidth]{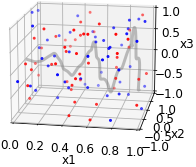}
\caption{Iteration 0 (Start)}
\end{subfigure}
\begin{subfigure}{0.3\linewidth}
\centering
\includegraphics[width=\linewidth]{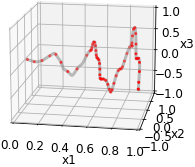}
\caption{Iteration 5 (Converged)}
\end{subfigure}
\caption{\small Case II: variable domain (SUHNPF). 
We restrict the four plots to three dimensions ($x_1$, $x_2$, and $x_3$) for visualization.}
\label{fig:illus-case3}
\end{figure}

This joint minimization case operates in a $n=30$ dimensional variable space. \textbf{Fig. \ref{fig:illus-case3}} shows the true Pareto front and SUHNPF convergence in the variable domain. Note the non-convexity in the variable domain, where $x_1$ varies uniformly between $[0,1]$, while $x_2,\ldots,x_{30}$ are sinusoidal in nature guided by $x_1$. Thus, the Pareto manifold has a spiral trajectory along $x_2,\ldots,x_{30}$ with evolution along $x_1$. 

Despite the Pareto front being convex, the objectives are non-convex. For MTL methods, the {\small \texttt{min\_norm\_solver}} \citep{sener2018multi}, which is integral to all MTL solvers, simply fails. Consequently, no MTL results are reported. 

For SUHNPF, following random initialization (iteration 0) in Fig. \ref{fig:illus-case3} (a), we observe that the candidate set $\mathcal{P}1$ propagates more towards increasing values of $x_1$ in Fig. \ref{fig:illus-case3}, and approximates the expected Pareto manifold at iteration 5. 

\begin{figure}[ht]
\centering
\begin{subfigure}{0.3\linewidth}
\centering
\includegraphics[width=\linewidth]{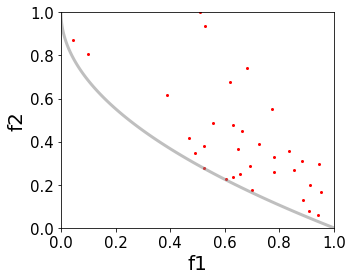}
\caption{Iteration 0 (Start)}
\end{subfigure}
\begin{subfigure}{0.3\linewidth}
\centering
\includegraphics[width=\linewidth]{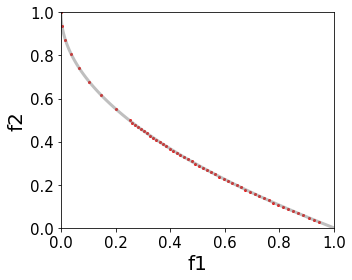}
\caption{Iteration 5 (Converged)}
\end{subfigure}
\caption{\small Case II: functional domain (SUHNPF).}
\label{fig:illus-form3}
\end{figure}


\subsection{Case III: \cite{tanaka1995ga}}

\resizebox{\linewidth}{!}{
\begin{minipage}{\linewidth}
\begin{align*}
&f_1(x_1,x_2) = x_1, \, f_2(x_1,x_2) = x_2\\
&\text{s.t.} \quad g_1(x_1,x_2)= (x_1-0.5)^2 + (x_2-0.5)^2 \leq 0.5\\
& g_2(x_1,x_2)= x_1^2 + x_2^2 - 1 - 0.1 \cos (16 \arctan ({x_1}/{x_2})) \geq 0\\
&g_3,g_4:0 \leq x_1, x_2 \leq \pi
\end{align*}
\end{minipage}
}

\begin{figure}[ht]
\centering
\begin{subfigure}{0.3\linewidth}
\centering
\includegraphics[width=\linewidth]{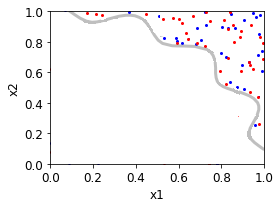}
\caption{Iteration 0 (Start)}
\end{subfigure}
\begin{subfigure}{0.3\linewidth}
\centering
\includegraphics[width=\linewidth]{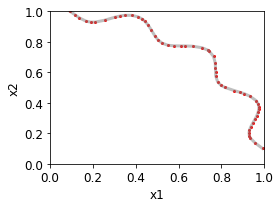}
\caption{Iteration 5 (Converged)}
\end{subfigure}
\caption{\small Case III: variable domain. The analytical solution for this problem is driven by constraints $g_1,g_2$. SUHNPF Pareto candidates $\mathcal{P}0$ (red dots) converge to the true front.}
\label{fig:illus-case2}
\end{figure}

For this joint minimization problem, the Pareto front is dominated by the two constraints $g_1$ and $g_2$, while linear functions $f_1$ and $f_2$ do not contribute to the Pareto optimal solution. \textbf{Fig. \ref{fig:illus-case2}} shows the convergence of SUHNPF Pareto candidates toward the known solution manifold. 

%


Because MTL approaches do not support constraints, they are not capable of solving this benchmark problem. However, note that if we were to remove constraints $g_1$ and $g_2$, $f_1$ and $f_2$ would then become independent of each other (and so not compete). The front then collapses to the point $(0,0)$, corresponding to the minimum of both functions. For this unconstrained problem, MTL methods would be expected to find this correct Pareto optimal solution point.

Case III highlights the need for any manifold based extractor to support both explicit and implicit forms of the Pareto front. Cases I and II have explicit form of front in the functional and variable domain. However, Case III has an implicit Pareto front (Fig. \ref{fig:illus-case2}) owing to constraints $g_1,g_2$, where they render an implicit relation between $x_1,x_2$ and therefore $f_1,f_2$. SUHNPF's ability to construct a full rank diffusive indicator function of Pareto \vs non-Pareto points enables it to approximate the true manifold.

\subsection{SUHNPF \vs OR and MTL Methods}

\textbf{Table \ref{tab:evalsOR}} reports the number of candidate evaluations by OR methods \vs SUHNPF to find $P=50$ Pareto points for Cases I-III. Because OR methods and SUHNPF all return $P$ true Pareto points, we compare methods on efficiency only. 

Note that HNPF is not an iterative solver: given a grid of points in the feasible domain, it identifies those that are weak Pareto optimal. 
It thus requires fewer evaluations with low variable dimensionality (Case I and III of 2 dimensions). In contrast, SUHNPF is a solver that starts with $50$ random points and iteratively converges them onto the weak Pareto front, irrespective of the variable space (Case II of 30 dimension), and hence is scalable to MTL problems. In general, the large number of evaluations required by OR methods is indicative of their lack of scalability to MTL problems. 

\begin{table}[bht]
\centering
\caption{\small The number of evaluations performed by each method to find 50 Pareto points across Cases I-III. HNPF performs well with small variable space dimensionality (\eg 2D in Case I \& III) but scales poorly to higher dimensionality (Case II, 30D).}
\begin{tabular}{c|r|r|r}
\toprule
\bf Method & {\bf Case I} & {\bf Case II} & {\bf Case III} \\
\midrule
NBI & 1,236,034 & 1,497,063,168 & 447,574 \\
mCHIM & 1,081,625 & 3,605,242,265 & 497,537 \\
PK & 138,224 & 169,223,715 & 377,410 \\ 
HNPF & {\bf 2,731} & 24,457 & {\bf 3,626} \\ \midrule 
\bf SUHNPF & 4,219 & {\bf 4,682} & 4,578 \\ \bottomrule  
\end{tabular}
\label{tab:evalsOR}
\end{table}


%

{\bf Table \ref{tab:evalsMTL}} reports the accuracy, efficiency and run-time of SUHNPF \vs MTL methods for Case~I. For Case II, the {\small \texttt{min\_norm\_solver}} \citep{sener2018multi} used by MTL methods fails, and Case III's constraints are not supported by MTL methods. Note that for fair evaluation, we only consider candidates that are produced within the feasible functional bounds for the problem. Additional run-time evaluation and discussion can be found in \textbf{Appendix \ref{app:runtime}}. 

\begin{table}[bht]
\centering
\caption{\small SUHNPF \vs MTL methods on Case I in finding $P=50$ Pareto points. We report the \% of feasible points each method finds and their avg/max error \vs the true front. Our error measure considers feasible points only; infeasible points are not penalized.}
\begin{tabular}{c|cccccc|c}
\toprule
\bf Method & LS & \!\!\!\!\!{\footnotesize MOOMTL}\!\!\!\!\! & PMTL\!\! & \!\!EPO\!\! & \!\!EPSE\!\! & \!\!PHN\!\! & \bf \!\!\!{\footnotesize SUHNPF}\!\!\! \\ \midrule 
Evaluations & 5K & 5K & 5K & 5K & 5K & 5K & 4,219 \\      
Run-time (secs) & 18.1 & 19.2 & 527 & 752 & 641 & 853 & 10.0 \\   
Points Found\!\! & 54\% & 32\% & 70\% & 68\% & 30\% & 80\% & 100\% \\
\!\!Avg Err ($10^{-4}$)\!\! & 0.53 & 0.45 & 4.15 & 8.73 & 0.61 & 3.04 & 0.52 \\
\!\!Max Err ($10^{-4}$)\!\! & 1.12 & 0.98 & 126 & 106 & 0.94 & 73.8 & 0.82 \\ 
\bottomrule   
\end{tabular}
\label{tab:evalsMTL}
\end{table}

Regarding Case I coverage and accuracy, SUHNPF returns all $50$ Pareto points; no MTL method does.  For all points that are found, we measure their error \vs the true Pareto front. SUHNPF is seen to achieve the lowest error, with maximum error bounded by the $10^{-4}$ error tolerance parameter set in our experiments. Specifically, the outer loop of Alg.~\ref{alg:fjc} would not achieve convergence until all the points points are within the prescribed error tolerance. In contrast, PMTL, EPO, and PHN yield maximum error two orders of magnitude larger. Note also that our error metric generously scores only the points found by each method, with no penalty for missing points. Visually, SUHNPF  (Fig. \ref{fig:illus-func}) clearly provides better coverage of the Pareto front via a denser, more even spread of points \vs those found by MTL methods (Fig.~\ref{fig:case1-mtl}).

Because MTL approaches assume convexity of objective functions to generate points with uniformity on the Pareto front, and Case I includes non-convex objectives, the MTL solvers fail to find points in certain regions (see Fig. \ref{fig:case1-mtl}). While EPO's solver has convergence criteria, it still produces points that did not converge (circled in blue). This stems from EPO's assumption on KKT conditions to achieve optimality, which fails on Case I's non-convex form of $f_2$. Correspondingly PHN(-EPO), which uses EPO as its base solver, also fails to converge on certain points. In contrast, 
SUHNPF relies on the FJC to test optimality, which fully supports non-convexity in functions and constraints. 

Regarding Case I efficiency, SUHNPF is also fastest: nearly twice as fast as LS and MOOMTL, more than 50x faster than PMTL and EPSE, 75x faster than EPO, and 85x faster than PHN. (Because PHN-EPO calls EPO, it is necessarily slower than EPO). As \cite{navon2021learning} note, LS is much faster than EPO, so one could expect PHN-LS to be faster than PHN-EPO and slower than LS.

\section{SUHNPF as a HyperNetwork}



Hypernetworks \citep{ha2016hypernetworks} train one neural model to generate effective weights for a second, target model. \citet{navon2021learning} and \citet{lin2021controllable} learn a neural manifold mapping MOO solutions to different target model weights, enabling the target model to achieve the desired Pareto trade-off for the MOO problem.


Assume the target task maps from input $Y$ to output $Z$. We seek to minimize objective functions $f_1$ and $f_2$ having loss functions $\mathcal{L}_1$ and $\mathcal{L}_2$. Given correct output $Z^*$, we score $Z$ for each loss function  $\mathcal{L}_i(Z,Z^*)$. A target model for this task $C_{\Theta}: Y \rightarrow Z$ with parameters $\Theta$ will yield loss $\forall_i \mathcal{L}_i(C_{\Theta}(Y),Z^*)$. The MOO problem is to find Pareto optimal $\Theta^*$ for the $f_1=\mathcal{L}_1$ \vs $f_2=\mathcal{L}_2$  trade-off. 

The objectives $\mathcal{L}_{1}(\Theta), \mathcal{L}_{2}(\Theta)$ for SUHNPF are continuous differentiable functions of $\Theta$. This enables SUNHPF's guided double gradient descent strategy to efficiently search the space of model target parameters $\Theta$, mapping each to resulting loss values $(\mathcal{L}_1,\mathcal{L}_2)$. Training data resulting from this search allows SUHNPF to learn an $\epsilon$-bounded approximation $\tilde{M}(\Theta^*)$ to the weak Pareto optimal manifold. 

As in prior Pareto Front Learning (PFL) work  \citep{navon2021learning,lin2021controllable}, this enables rapid model personalization at run-time based on user preferences.  The neural MOO 
Loss$_{classifier}$ is a weighted linear combination of the user-prescribed objectives ($\mathcal{L}_1,\mathcal{L}_2$). The classifier loss hyper-parameter $\alpha$ (trade-off value) is computed as a post-processing step corresponding to Pareto optimal classifier weights $\Theta^*$ for rapid traversal of arbitrary $(\alpha, \Theta^*)$ solutions. See \textbf{Fig. \ref{fig:framework}} in \textbf{Appendix \ref{app:hypernet}} for additional details of the setup of SUHNPF as a hypernetwork to optimize a target model. 
\begin{figure*}[ht]
\centering
\begin{subfigure}{0.3\linewidth}
\centering
\includegraphics[width=\linewidth]{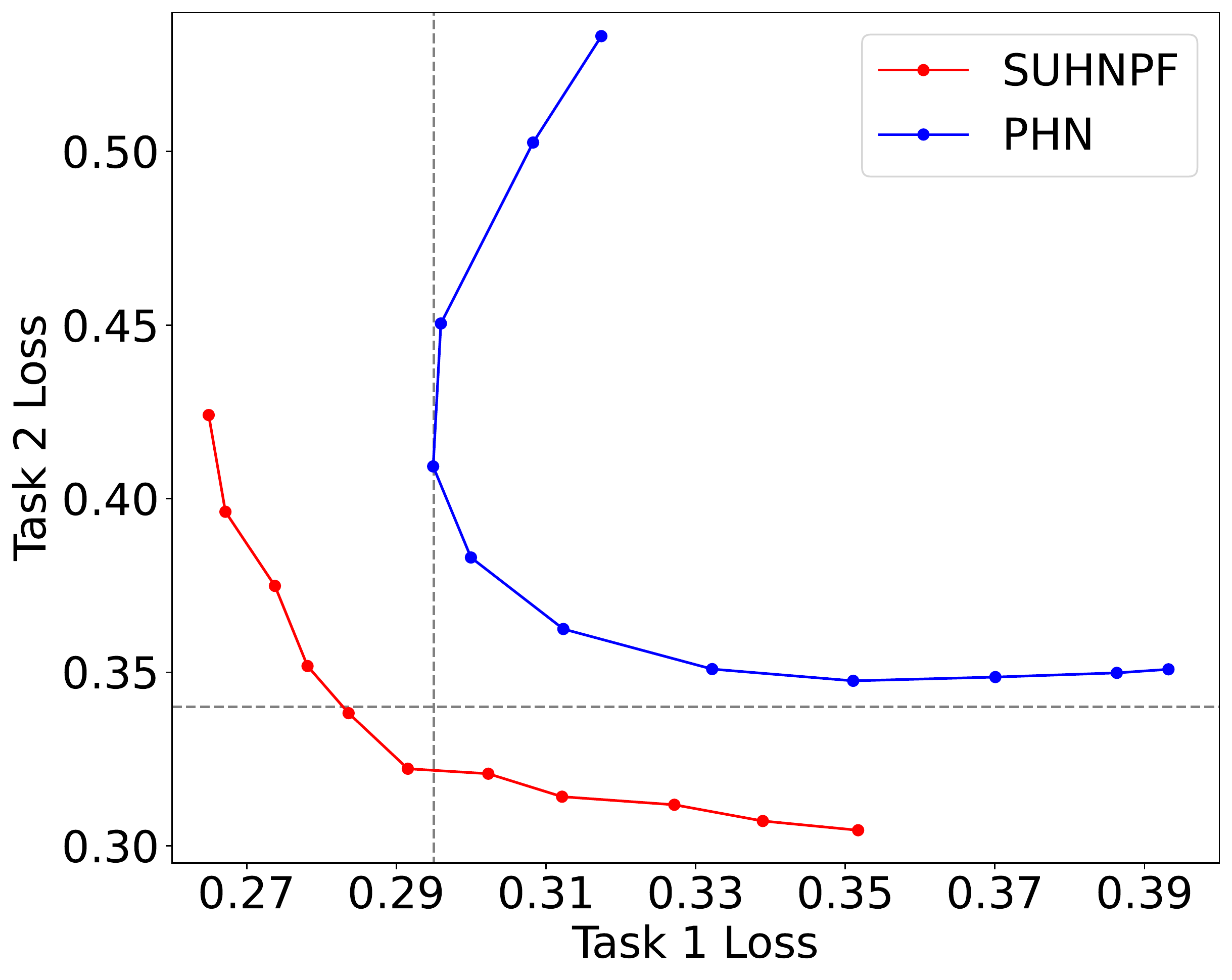}
\caption{\scriptsize MultiMNIST}
\end{subfigure}
\quad
\begin{subfigure}{0.3\linewidth}
\centering
\includegraphics[width=\linewidth]{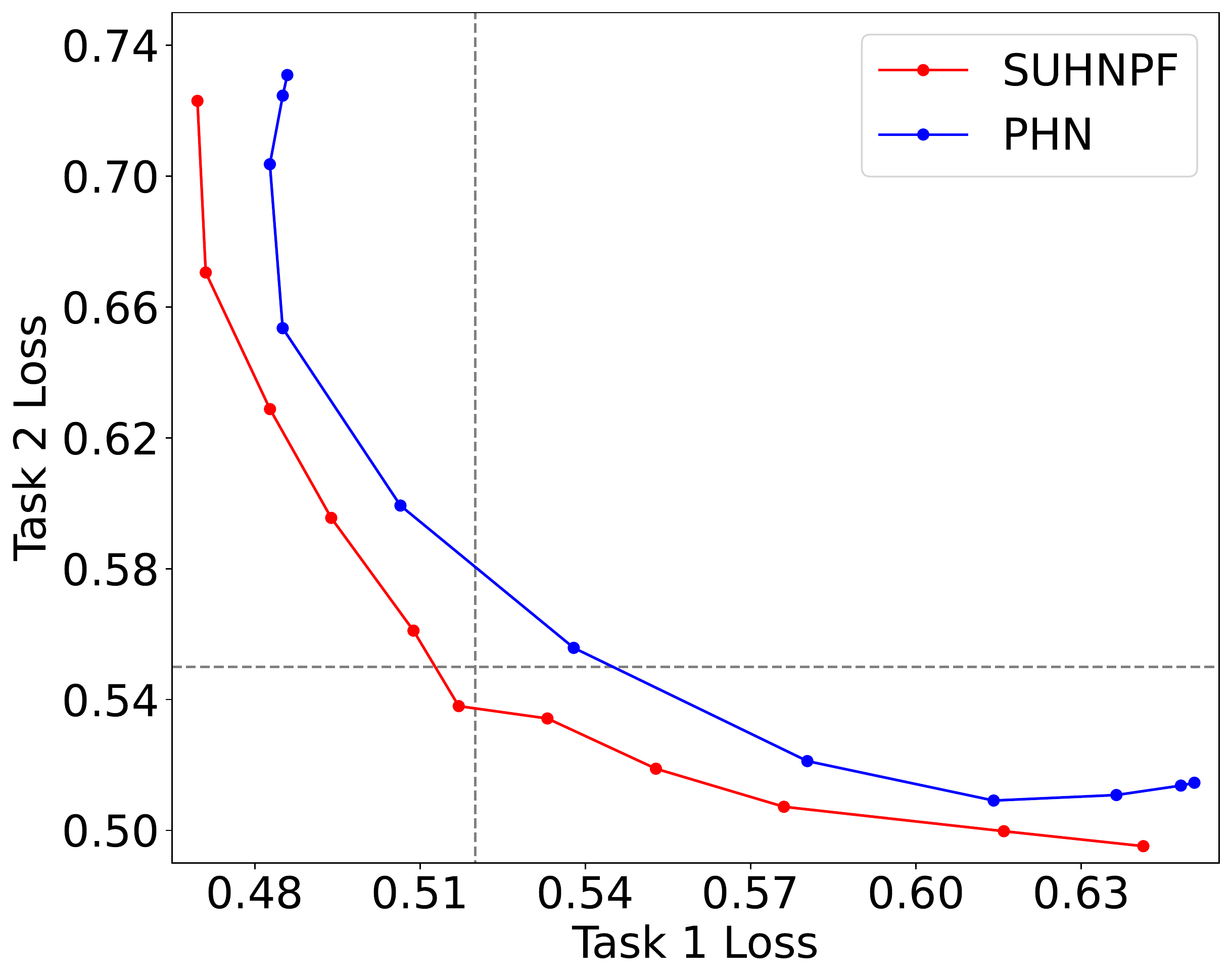}
\caption{\scriptsize MultiFashion}
\end{subfigure}
\quad
\begin{subfigure}{0.3\linewidth}
\centering
\includegraphics[width=\linewidth]{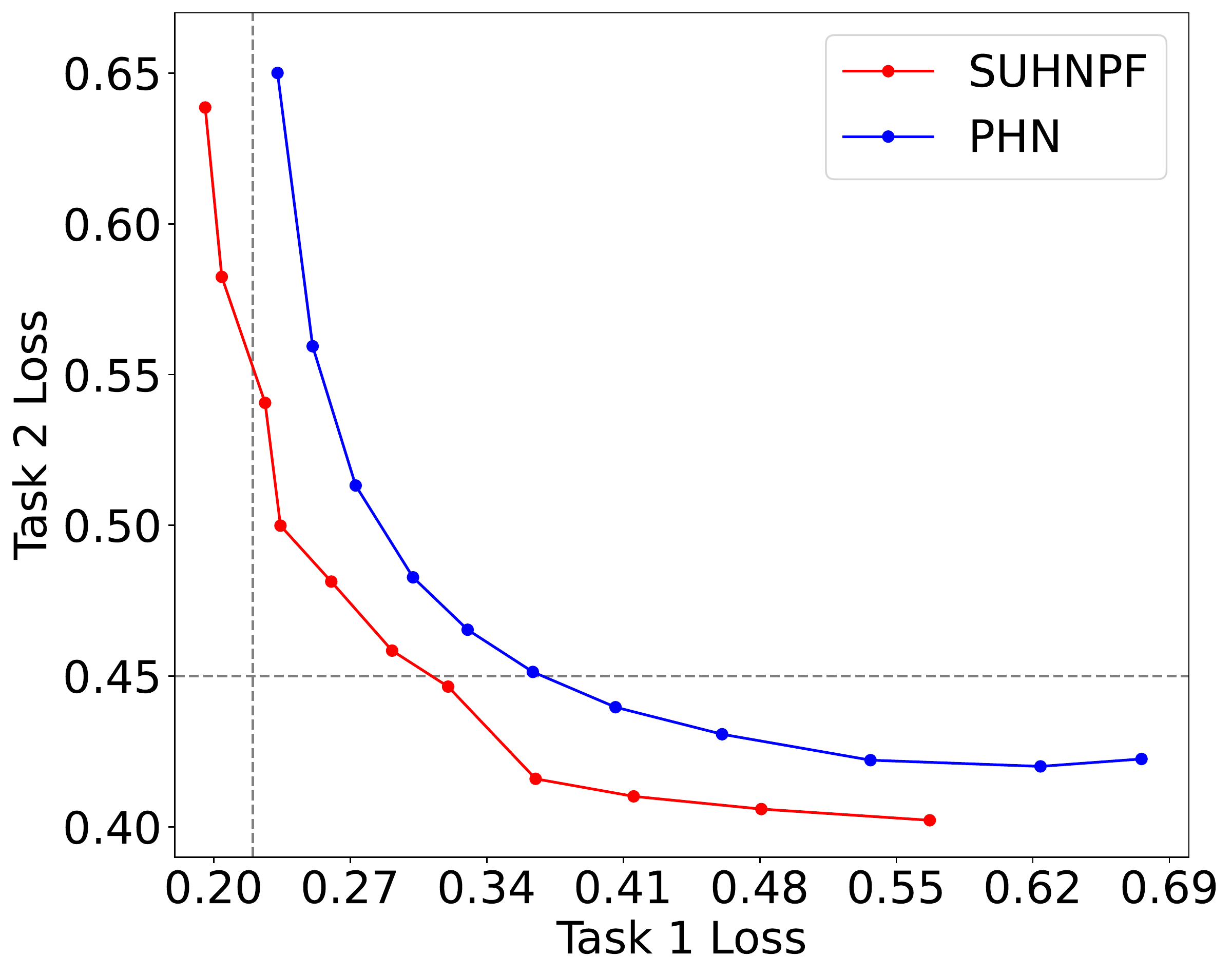}
\caption{\scriptsize MultiFashion+MNIST}
\end{subfigure}
\caption{\small Cross-entropy loss on the test split for all three MTL datasets for SUHNPF \vs PHN. The 11 points shown for each method correspond (from left-to-right) to varying trade-offs preferences in minimizing the combined linear loss over objectives: $\alpha f_1 + (1-\alpha) f_2$ \,for\, $\alpha \in \{1, \,0.9, \ldots,\, 0\}$. The gray dashed-line show the best loss achieved by LeNet to classify a single image for each given task.
}
\label{fig:mnist_err}
\end{figure*}

\subsection{Evaluation on Multi-Task Learning}

We evaluate on the same MTL image classification problems as in \citet{navon2021learning}. Given two underlying source datasets, MNIST \citep{lecun1998gradient} and Fashion-MNIST \citep{xiao2017fashion}, \citet{navon2021learning} report on three MTL tasks: MultiMNIST \citep{sabour2017dynamic}, Multi-Fashion, and Multi-Fashion + MNIST. In each case, two images are sampled from source datasets and overlaid, one at the top-left corner and one at the bottom-right, with each also shifted up to 4 pixels in each direction. The two competing tasks are to correctly classify each of the original images: Top-Left (Task 1 or $f_1$) and Bottom-Right (Task 2 or $f_2$). We use $120$K training and $20k$ testing examples and directly apply existing single-task models, allocating $10\%$ of each training set for constructing validation sets, as used in \cite{lin2019pareto}. \cite{navon2021learning} found that PHN-EPO (henceforth PHN) was more accurate than other methods they compared, so we use PHN as our baseline.

We adopt the LeNet architecture \citep{lecun1998gradient} as the target model to learn. Following prior MTL work \citep{sener2018multi}, we treat all layers other than the last as the shared representation function and put two fully-connected layers as task-specific functions. We use cross-entropy loss with softmax activation for both task-specific loss functions. Because cross-entropy loss functions are differentiable, we can use them directly as training objectives.



{\bf Results.} We see SUHNPF \vs PHN results on dataset test splits in \textbf{Fig. \ref{fig:mnist_err}}. Because SUHNPF defines a strict $\epsilon$-bound on error, we can assert its correctness on this basis alone. Visual inspection also shows that PHN returns dominated points (\eg top of MultiMNIST plot), whereas a Pareto front by definition includes only non-dominated points.  Nonetheless, we cannot directly measure error \vs a known Pareto front because real MOO problems lack a simple analytical solution like synthetic benchmark problems. Of course, we can still compare relative performance of methods. We see that {\em SUHNPF achieves strictly lower loss than PHN across all user trade-off settings of $\alpha$ on all three datasets}. 

Since the minimum loss $\textnormal{min}(f_1)$=$\textnormal{min}(f_2)$=0, for both objectives, the ideal point \citep{marler2004survey} for joint minimization is $(0,0)$. A simple error measure for each point found is thus its $L2$ distance from $(0,0)$: $\sqrt{f_1^2 + f_2^2}$. \textbf{Table~\ref{tab:mtlalphas}} reports this distance for each Pareto point found at each $\alpha$ (across methods and datasets). We also report the average over the 11 settings of $\alpha$. Overall, Table \ref{tab:mtlalphas} quantifies what Fig. \ref{fig:mnist_err} depicts visually: SUHNPF performs strictly better for every Pareto point and thus also on average. 

\begin{table}[ht]
\centering
\caption{\small SUHNPF \vs PHN on MTL tasks, measured by distance of each Pareto point found \vs the ideal loss point $(f_1,f_2)=(0,0)$.}
\begin{tabular}{c|ccccccccccc|c} \toprule
& \multicolumn{12}{c}{Trade-off values $\alpha$}\\ 
\!\!Method\!\! & 0.0 & 0.1 & 0.2 & 0.3 & 0.4 & 0.5 & 0.6 & 0.7 & 0.8 & 0.9 & 1.0 & {\bf Avg} \\ \midrule
\multicolumn{13}{c}{MultiMNIST}\\ 
PHN & .621 & .585 & .539 & .504 & .486 & .478 & .483 & .494 & .508 & .521 & .527 & {\bf .522} \\
\!\!\!\!\!{\footnotesize SUHNPF}\!\!\!\!\! & .500 & .478 & .464 & .448 & .441 & .434 & .441 & .443 & .452 & .457 & .465 & {\bf .456} \\ \midrule
\multicolumn{13}{c}{MultiFashion}\\ 
PHN & .877 & .872 & .853 & .813 & .784 & .773 & .779 & .797 & .816 & .826 & .829 & {\bf .819} \\
\!\!\!\!\!{\footnotesize SUHNPF}\!\!\!\!\! & .862 & .819 & .792 & .773 & .757 & .746 & .754 & .758 & .767 & .793 & .810 & {\bf .784} \\ \midrule
\multicolumn{13}{c}{MultiFashion+MNIST}\\ 
PHN & .690 & .613 & .581 & .569 & .571 & .579 & .598 & .631 & .682 & .752 & .797 & {\bf .642} \\
\!\!\!\!\!{\footnotesize SUHNPF}\!\!\!\!\! & .667 & .617 & .586 & .552 & .547 & .543 & .549 & .553 & .583 & .629 & .695 & {\bf .593} \\ \bottomrule
\end{tabular}
\label{tab:mtlalphas}
\end{table}

\section{Understanding SUHNPF \vs PHN} \label{sec:comp}
\begin{figure}[ht]
\centering
\begin{subfigure}{0.3\linewidth}
\centering
\includegraphics[width=\linewidth]{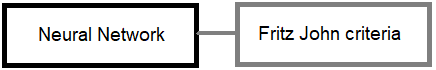}
\caption{SUHNPF}
\end{subfigure}
\begin{subfigure}{0.3\linewidth}
\centering
\includegraphics[width=\linewidth]{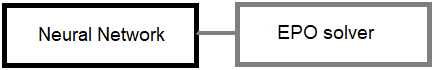}
\caption{PHN}
\end{subfigure}
\caption{\small High level abstraction of SUHNPF and PHN solvers. While SUHNPF uses the Fritz-John criteria for optimality check, PHN uses the candidates deemed optimal by the EPO solver. 
}
\label{fig:hnpfvsphn}
\end{figure}

While both SUHNPF and PHN 
are manifold-based (\textbf{Fig. \ref{fig:hnpfvsphn}}), they differ in the type of manifold being learned. SUHNPF explicity maintains point sets $\mathcal{P}0$ and $\mathcal{P}1$ to learn the classification boundary between Pareto \vs non-Pareto points as per the FJC. PHN fits a regression surface over the set of points returned by LS or EPO. Since neither LS nor EPO are guaranteed to operate under non-convex settings (Section \ref{sec:related}), those drawbacks are in turn inherited by PHN in using them. \textbf{Table \ref{tab:hnpfvsphn}} highlights the key differences. The distinction between a diffusive full-rank indicator \vs a low-rank regressor is further discussed in \textbf{Appendix~\ref{app:comp}}. 


\begin{table}[ht]
\centering
\caption{\small SUHNPF \vs PHN for Pareto front learning.}
\begin{tabular}{l|c|c}
\toprule
\bf Criteria & \textbf{SUHNPF} & \textbf{PHN} \\
\midrule
Handle non-convexity\!\! & \cmark & \xmark \\
Supports constraints & \cmark & \xmark \\
Manifold Extractor & \cmark & \cmark \\
Nature of manifold & Diffusive full-rank indicator & Low-rank regressor\!\! \\
Optimality Criteria & Fritz-John Conditions  & EPO solver \\
\bottomrule
\end{tabular}
\label{tab:hnpfvsphn}
\end{table}

\section{Conclusion}

Multi-objective optimization problems require balancing competing objectives, often under constraints. In this work, we described a novel method for {\em Pareto-front learning} (inducing the full Pareto manifold at train-time so users can pick any desired optimal trade-off point at run-time). Our SUHNPF Pareto solver is robust against non-convexity, with error bounded by a user-specified tolerance. Our key innovation over prior work's HNPF \citep{singh2021hybrid} is to exploit Fritz-John Conditions for a novel guided {\em double gradient descent} strategy. The scaling property imparts significant improvement in memory and run-time \vs prior OR and Multi-Task Learning (MTL) approaches. Results across synthetic benchmarks and MTL problems in image classification show clear, consistent advantages of SUHNPF in capability (handling non-convexity and constraints), denser coverage and higher accuracy in recovering the true Pareto front, and efficiency (time and space). Beyond empirical results, our conceptual framing and review of prior work also further bridges disparate lines of OR and MTL research.

Both SUHNPF and MTL methods assume differentiable evaluation metrics as training loss so optima to be found through gradient descent. However, loss can be 
a non-differentiable, probabilistic measure, such as in fairness-related tasks \citep{sacharidis2019top,valdivia2020fair}. This creates a risk of metric divergence between training loss \vs the evaluation measure of interest \citep{abou2012note}. Continuing development of differentiable measures can help to address this \citep{swezey2021pirank}.

\clearpage

\bibliographystyle{ACM-Reference-Format}
\bibliography{References}

\appendix

\section{SUHNPF as HyperNetwork} \label{app:hypernet}
\vspace{-1em}

\begin{figure}[h]
	\centering
	\includegraphics[width=0.5\linewidth]{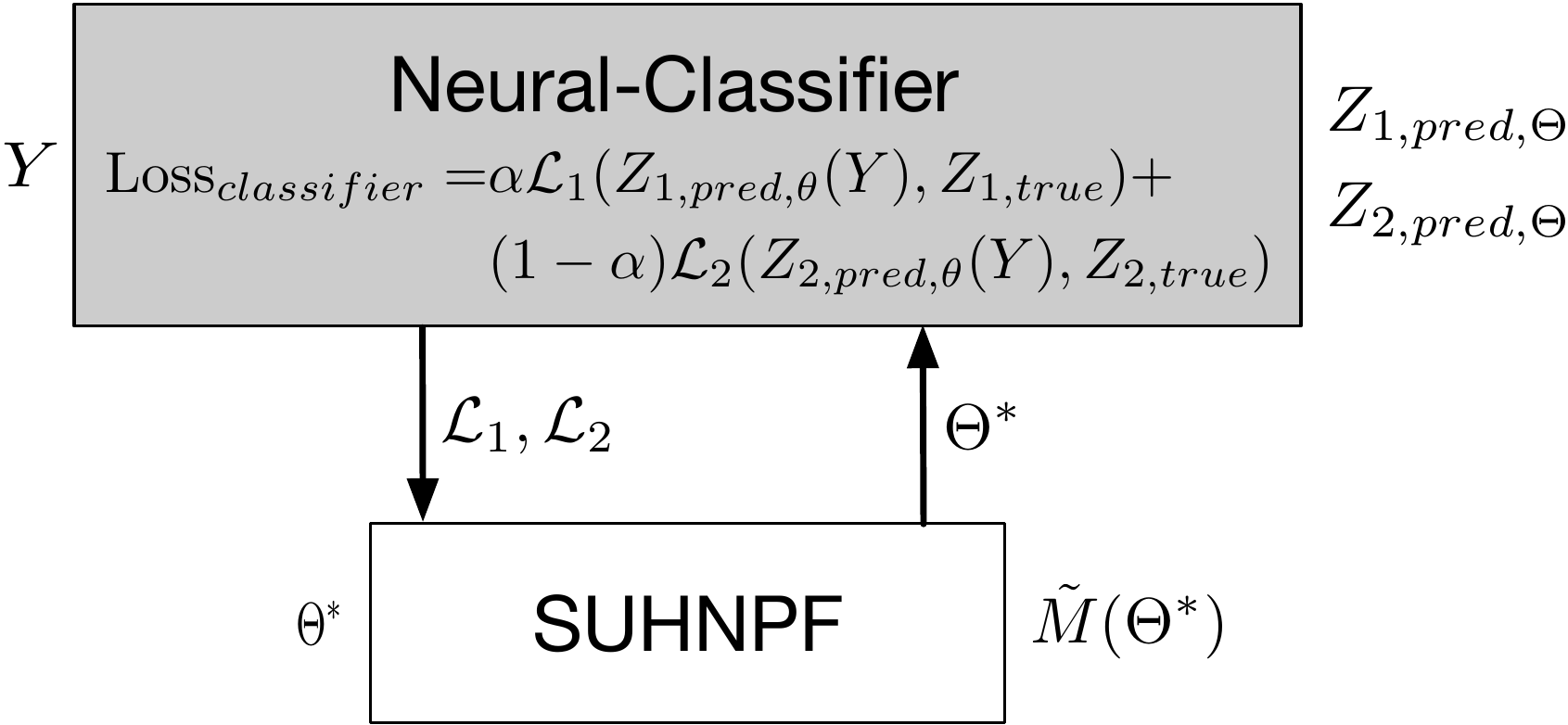}
	\caption{\small Framework for extracting the Pareto optimal front $\tilde{M}(\Theta)$ of a given target model $C_{\Theta}$ (which could also be a non-neural model: Decision Tree, Logistic Regression, \textit{etc}.)}
	\label{fig:framework}
\end{figure}

\textbf{Fig. \ref{fig:framework}} shows the overview of SUHNPF as a hypernetwork tasked with optimizing the weights of the target neural classifier. The input to the neural classifier are data points $Y$ and the output are matched against labels $Z_{1,true},Z_{2,true}$ for two different tasks. The weights of the neural classifier are $\Theta$ and SUHNPF as a hypernetwork approximates the weak Pareto manifold $\tilde{M}(\theta^*)$ for optimal trade-off over different values $\alpha$ for the two MOO losses $\mathcal{L}_1,\mathcal{L}_2$.

\vspace{-1em}
\section{Space Complexity Analysis} \label{app:space}
\vspace{-1em}

MTL methods solve problems in both primal and dual space \ie gradient of the objectives in the primal and the trade-off $\alpha$'s in the dual. SUHNPF however works only in the primal space \wrt the gradient of the functions necessary in the construction of the Fritz-John matrix, since the FJC ensure $\alpha$ free stationary point identification. Thus, the additional dual optimization space is not required. To fairly compare \wrt MTL methods, we consider the general cost of both such systems under a unconstrained setting \ie only objectives and no additional constraints. Thus $k,n$ indicate the number of objectives and the dimension of the variable space.

\textbf{SUNHPF}. To find $P$ Pareto candidates, SUHNPF updates $P$ points of size $Pn$. The $\nabla F^T \nabla F$ and $\nabla det$ matrices are of size $k^2$ and $nk$ respectively. The total memory cost is thus of order $O(n(P+k)+k^2))$.

\textbf{MTL}. To find $P$ Pareto candidates, MTL methods uses $P$ cones or rays requiring size $Pn$. The gradient matrix of the objective function $\nabla F$ takes $nk$, constructing the simplex takes $k^2$, solving for trade-off $\alpha$ takes $k^2$ and the iterative update requires additional $nk$ memory. The total memory cost is thus of order $O(n(P+2k)+2k^2))$.

\vspace{-1em}
\section{Run-time Analysis}\label{app:runtime}
\vspace{-1em}

While correctness and point density in finding the true Pareto optimal solution should be our top priority in comparing methods, we also report run-time of SUHNPF \vs other MTL approaches on the studied cases. As in Table~\ref{tab:evalsMTL}, we explicitly request that each method generate $50$ Pareto candidates, within the feasible functional domain. \textbf{Table~\ref{tab:evalsMTL}} reports the overall execution time, averaged over $10$ runs each, given our experimental setup in  Appendix \ref{app:setup}.


PHN uses either EPO or LS as their base solver, hence we report the total time that includes the (a) run-time of the base solver; and (b) the neural network run-time to learn the regression manifold. Cases I and III have a 2D variable domain, where SUHNPF takes $1$s  per epoch, with $2$ epochs for training in Step 7 of Alg. \ref{alg:fjc}. Both the cases took $5$ epochs to converge, resulting in a total run-time of $10$s. Case II has a 30D variable domain where SUHNPF takes $2$s per epoch resulting in a total run-time of $20$s. While LS and MOOMTL are at similar run-time scale with SUHNPF, they fail to generate an even spread of points (Fig. \ref{fig:case1-mtl} (a,b)).

\section{SUHNPF \vs Other MTL methods} \label{app:comp}

\textbf{Point based solvers}. Most MTL methods, including 
MOOMTL, 
PMTL, 
EPSE, 
and EPO 
are point based solvers. Being point-based, they return one solution per run, relying upon specialized local initialization to generate an even spread of Pareto points, using cones, rays, or other domain partitioning strategies, across the feasible set of saddle points. Thus asked for $P$ Pareto candidates, these solvers would have to run for $P$ instances. Later, if the user demands $2P$ points, they have to run for $2P$ instances from scratch, without utilizing the results from the previous run.

\textbf{Manifold based solvers}. A manifold-based solution strategy should separate Pareto \vs non-Pareto points, without requiring any special initialization. It would also be able to extract $2P$ Pareto candidates while being trained to generate $P$ candidates, due to interpolating from the learned boundary. This is highly advantageous over point-based schemes for deployment of practical systems, where the expected user trade-off preference is not known \textit{a priori}, hence good to have the full approximated front. Notably, \citet{navon2021learning} and \citet{lin2021controllable} are the only prior manifold based Pareto solvers that we are aware of that are also scalable to optimize large neural models. Another advantage is that both SUHNPF and EPO (used by PHN in backend) solvers have a user-specified error tolerance criteria built in, while other MTL solver lack it and therefore run a specified number of iterations before declaring a candidate Pareto, without actually checking for optimality. 

\textbf{Full rank indicator \vs low rank regressor}. A manifold based solver should also generalize to cases where the manifold is an implicit function as opposed to its easier counterpart of being an explicit function. SUHNPF has an added advantage in extracting the weak Pareto manifold as an $k$-dimensional diffusive indicator function as opposed to a $(k-1)$-dimensional manifold itself, where the regressed manifold is not only guided by the weak Pareto points (indicator value 1) but also the sub-optimal points (indicator value 0) for a more robust and accurate extraction. Thus it can generally approximate the manifold, irrespective of the manifold being an explicit or implicit function. In comparison, PHN learns a $(k-1)$-dimensional regression manifold, given solution points obtained from EPO or LS. Therefore, PHN's default assumption is that the Pareto manifold is always an explicit function \ie for $k$ objectives, the Pareto manifold is of dimension $k-1$.

\providecommand{\upGamma}{\Gamma}
\providecommand{\uppi}{\pi}

\section{Discussion on Remark 1}\label{app:fjc}

\textbf{Remark}: \small \textit{If $f_i$s are continuous and differentiable once, in an unconstrained setting, then the set of weak pareto optimal points are $x^*=\{x|det(L(x)^TL(x))=0\}$, for a non-square matrix $L(x)$, and is equivalent to $x^*=\{x|det(L(x))=0\}$ for a square matrix $L(x)$.}

\subsection{Illustration}

We begin by considering two multi-variable functions (for ease of description). Let us consider the following two quadratic functions, convex in both variables as:
\begin{align*}
	f_{1}(\mathbf{x}) = (x_{1}-1)^{2} + (x_{2}-1)^2\\
	f_{2}(\mathbf{x}) = (x_{1}+1)^2 + (x_{2}+1)^2
\end{align*}

\begin{figure}[ht]
	\centering
	\includegraphics[width=0.3\linewidth]{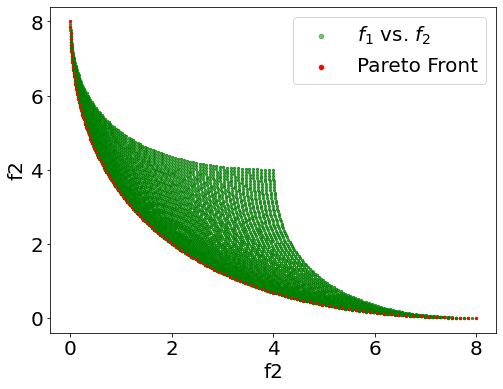}
	\caption{\small Functional Domain plot for two competing objectives.}
	\label{fig:parabola}
\end{figure}

The task is to find the Pareto front between the two objectives $f_{1}(\mathbf{x})$ and $f_{2}(\mathbf{x})$. Since this is a trivial problem the Pareto front is known a-priori as the straight line $x_{1} = x_{2}$ for $x_{1} \in [-1,1]$ in the variable domain. Let us now first plot $f_{1}$ \vs $f_{2}$ for visual assessment in \textbf{Fig. \ref{fig:parabola}}.

Note that independent of each other:
\begin{align*}
	\nabla f_{1}(\mathbf{x})& = \left[\frac{\partial f_{1}}{\partial x_{1}} \quad \frac{\partial f_{1}}{\partial x_{2}}\right]^{T} = \mathbf{0} \,\, \mathrm{at} \,\, (x_{1},x_{2}) = (1,1)\\
	\nabla f_{2}(\mathbf{x}) &= \left[\frac{\partial f_{2}}{\partial x_{1}} \quad \frac{\partial f_{2}}{\partial x_{2}}\right]^{T} = \mathbf{0} \,\, \mathrm{at} \,\, (x_{1},x_{2}) = (-1,-1)
\end{align*}

One can easily confirm that the gradient matrix $L$ cannot be identically zero for any value of $x \in \mathbb{R}^{2}$
\begin{align*}
	L = [\nabla f_{1}(x) \quad \nabla f_{2}(x)]
\end{align*}

Note that in the above Fig. \ref{fig:parabola} we plotted the analytical solution explicitly. However for a Pareto solver we need to find the red curve for any two functions $f_{1}$ and $f_{2}$.

\textbf{Target}: Visually, we want an algorithm with an arbitrary initialization of a point $(x_{1},x_{2})$ such that ($f_{1},f_{2}$) is in the feasible region (shaded green above) to converge (or terminate) at the Pareto front (red curve) that satisfies $x_{1} = x_{2}$ with a user prescribed tolerance.

Let us say we have an initialization $\mathbf{x} = [x_{1} \quad x_{2}]$ at iterate $i=0$. The iterative update can then be written as:
\begin{align*}
	\mathbf{x}^{i+1} = \mathbf{x}^{i} \pm \eta Q(\mathbf{x})
\end{align*}
where $\pm$ is used to indicate that a choice of minimization or maximization is not yet decided.

\textbf{Requirement 1}: Here, $\eta > 0$ is the step size (user-prescribed) and the unknown vector function $Q(\mathbf{x})$ should be such that $Q(\mathbf{x}) = \mathbf{0}$ (termination) when $x_{1} = x_{2}$ (Pareto front). Otherwise, the iteration will not terminate and can over/undershoot the Pareto front which we already know as an analytical solution for the above problem.

\textbf{Requirement 2}: The vector function $Q(\mathbf{x})$ should be somehow related to the original objectives $f_{1}$ and $f_{2}$.


\subsection{Choice of $Q(\mathbf{x})$ and Convergence}
The most common (and easy to prescribe) choice of $Q(\mathbf{x})$ is given by a linear scalarization of the original objectives $f_{1}$ and $f_{2}$ that linearly combines these two (or more) objectives to create a single scalar objective function:
\begin{align}
	S(\mathbf{x}) = \alpha_{1} f_{1}(x) + \alpha_{2} f_{2}(x) \label{eq:linscal}
\end{align}

The vector update function $Q(\mathbf{x})$ is now given by the gradient of $S(\mathbf{x})$ with respect to variables $x_{1}$ and $x_{2}$, (given $\alpha_{i}$s):
\begin{align*}
	Q(\mathbf{x}) = L\alpha = [\nabla f_{1}(x) \quad \nabla f_{2}(x)] [\alpha_{1} \quad \alpha_{2}]^{T}
\end{align*}

\subsection{Derivation of Fritz-John Conditions}

We already know that $L$ or the gradient matrix cannot be identically zeros as discussed before. Furthermore, to avoid a trivial solution the vector $[\alpha_{1} \quad \alpha_{2}]$ must also not be identically zero. This becomes clear if the scalarized function $S(\mathbf{x})$ is defined.

The only remaining possibility is $L\alpha$ should approach zero for some $\mathbf{x} = [x_{1} \quad x_{2}]$ as we iteratively update $x$ using $Q(\mathbf{x})$. This gives us our termination/convergence criterion. Now let us look at what does $L\alpha = 0$ imply to understand the role of $\alpha$.
\begin{align*}
	L\alpha &= [\nabla f_{1}(x) \quad \nabla f_{2}(x)] _{2}[\alpha_{1} \quad \alpha_{2}]^{T} \\
	&= \left[\begin{matrix} \frac{\partial f_{1}}{\partial x_{1}} & \frac{\partial f_{2}}{\partial x_{1}}\\ \frac{\partial f_{1}}{\partial x_{2}} & \frac{\partial f_{2}}{\partial x_{2}}\end{matrix}\right] [\alpha_{1} \quad \alpha_{2}]^{T} = [0 \quad 0]^{T}
\end{align*}

Let us assume any point $(x_{1},x_{2})$ in the feasible domain. What $\alpha$ values can achieve the above termination criterion? We now have two equations in two unknowns ($\alpha_{i}s$):
\begin{align*}
	\alpha_{1}\frac{\partial f_{1}}{\partial x_{1}} + \alpha_{2} \frac{\partial f_{2}}{\partial x_{1}} = 0 \\
	\alpha_{1}\frac{\partial f_{1}}{\partial x_{2}} + \alpha_{2} \frac{\partial f_{2}}{\partial x_{2}} = 0
\end{align*}

Eliminating $\alpha_{1}$ using the first equation and substituting in the second equation:
\begin{align*}
	\left[-\left(\frac{\partial f_{1}}{\partial x_{2}} \frac{\partial f_{2}}{\partial x_{1}}\right)/\left(\frac{\partial f_{1}}{\partial x_{1}} \right) + \frac{\partial f_{2}}{\partial x_{2}}\right] \alpha_{2} = 0
\end{align*}

For any $\alpha_{2} >0$, this implies: 
\begin{align*}
	&\left[-\left(\frac{\partial f_{1}}{\partial x_{2}} \frac{\partial f_{2}}{\partial x_{1}}\right)/\left(\frac{\partial f_{1}}{\partial x_{1}} \right) + \frac{\partial f_{2}}{\partial x_{2}}\right] = 0\\
	\Rightarrow &\left[ \frac{\partial f_{1}}{\partial x_{1}} \frac{\partial f_{2}}{\partial x_{2}} - \frac{\partial f_{1}}{\partial x_{2}} \frac{\partial f_{2}}{\partial x_{1}} \right] = 0
\end{align*}

Alternatively,
\begin{align*}
	det \left(\left[\begin{matrix} \frac{\partial f_{1}}{\partial x_{1}} & \frac{\partial f_{2}}{\partial x_{1}}\\ \frac{\partial f_{1}}{\partial x_{2}} & \frac{\partial f_{2}}{\partial x_{2}}\end{matrix}\right] \right) = det(L)= 0
\end{align*}

Note that for any matrix $A \neq \mathbf{0}$, $Ax=0$ can be solved for a non-trivial $x \neq 0$ if and only if A has a null-space; or A is low rank; or if A is square then it's determinant is zero.

For the two convex objectives $f_{1}$ and $f_{2}$ described in the beginning, the user can trivially evaluate $det(L) = 0$ to arrive at the analytical solution for the Pareto front ($x_{1} = x_{2}$) independent of the choice of the trade-off values $\alpha_{i}$s. 

In effect, this implies that any stationary point at the which the above iteration will stop corresponds to an $(x_{1},x_{2})$ such that matrix $L$ becomes low rank. This is what the Fritz-John criteria indicates directly. Note that the Fritz-John conditions are generalized to any form of objectives, be they convex or non-convex. As long as the system $L(x)$ is rank-deficient \ie $det(L(x))=0$, we are certain to conclude that the point $x$ in the feasible set is indeed a weak Pareto point of the system. 

\subsection{Extension to non-square systems}

The $det(L)$ matrix defined in Eq. \ref{eq:fjmatrix} is given by:
\begin{align*}
	L = \begin{bmatrix}
		\nabla F & \nabla G \\
		\mathbf{0} & G
	\end{bmatrix}
\end{align*}
To achieve $det(L)=0$ requires that either:

\begin{enumerate}[leftmargin=*]
	\item $\nabla F(x)=0$: atleast one objective function has reached its optimum (local/global minima/maxima under a min/max setting); {\em and / or} 
	\item  $G(x)=0$: at least one constraint is satisfied.
\end{enumerate}
This criteria is only applicable for square systems. However, for practical problems, the system might become non-square, hence we need to satisfy $det(L^TL)=0$ following Eq. \ref{eq:paropt}. One might think that it's a different optimization problem. However satisfying $det(L^TL)=0$ mathematically provides the same justification and we provide the derivation of it.
\begin{align*}
	det(L^TL) &= \begin{bmatrix}
		\nabla F^T & \mathbf{0} \\
		\nabla G^T & G^T
	\end{bmatrix} \begin{bmatrix}
		\nabla F & \nabla G \\
		\mathbf{0} & G
	\end{bmatrix}\\
	&= \begin{bmatrix}
		\nabla F^T \nabla F & \nabla F^T \nabla G \\
		\nabla G^T \nabla F & \nabla G^T \nabla G + G^TG
	\end{bmatrix} \numberthis \label{eq:efjc}
\end{align*}
We now observe Eq. \ref{eq:efjc} for the two cases prescribed above and see if $det(L^TL)$ evaluates to zero or not. For Case 1, where $\nabla F=0$, Eq. \ref{eq:efjc} reduces to:
\begin{align*}
	det(L^TL) &= \begin{bmatrix}
		\mathbf{0} & \mathbf{0} \nabla G \\
		\nabla G^T \mathbf{0} & \nabla G^T \nabla G + G^TG
	\end{bmatrix}
\end{align*}
which is low-rank since row 1 equates to 0.
For Case 2, where $G=0$, Eq. \ref{eq:efjc} reduces to:
\begin{align*}
	det(L^TL) &= \begin{bmatrix}
		\nabla F^T \nabla F & \nabla F^T \nabla G \\
		\nabla G^T \nabla F & \nabla G^T \nabla G + 0
	\end{bmatrix}\\
	&= \nabla F^T \nabla G^T \begin{bmatrix}
		\nabla F & \nabla G \\
		\nabla F & \nabla G
	\end{bmatrix}
\end{align*}
which is low-rank again because row 1 and row 2 are equal. Hence it is easy to observe that satisfying $det(L)=0$ is equivalent to satisfying $det(L^TL)=0$.

\subsection{Effect of Trade-off on Optimality} \label{app:remark1}

Let us consider a general gradient matrix $L$ now.

\textbf{Case 1}: If the gradient matrix $L$ is full rank. Then $\alpha = \mathbf{0}$  (vector is identically zero) is the only solution.

\textbf{Case 2}: If the gradient matrix $L$ has a rank deficiency $q=1$ (one rank deficient). Then one of the $\alpha_{i}$s can be chosen arbitrarily or only one equation is missing. A simplex criterion $\sum_{i}\alpha_{i} = 1$ then supplies this arbitray choice. Note that one could also choose $\sum_{i}\alpha_{i}^{2} = constant$ or any other such choice to make $\alpha_{i}$ values uniquely determinable. For convex objectives $f_{1}$ and $f_{2}$ as described in the beginning, the reader can easily check that the $L$ matrix is only one rank deficient for all $\mathbf{x} \in \mathbb{R}^{2}$.

\textbf{Case 3}: \textbf{Here comes the trouble}. If matrix $L$ has a rank deficiency $q>1$ (more than one rank deficient). Now more than one of the $\alpha_{i}$s can be chosen arbitrarily. The simplex criterion is now no longer sufficient to determine a unique $\alpha$ vector. This is the most general case for an arbitrary number of non-convex objective functions $f_{1},\cdots,f_{k}$. Finding the Pareto optimal front would first require us to resolve the rank of $L$. Without this an adaptive update on $\alpha_{i}$ has no bearing whatsoever.

Here the Fritz John necessary conditions \ie $det(L) = 0$ is the most general way to state rank deficiency of $L$ irrespective of whether the matrix is one or more than one rank deficient.





\section{Experimental Setup Details}
\label{app:setup}
\label{appendix:termination}

{\bf Experimental Setup}. We use an Nvidia 2060 RTX Super 8GB GPU, Intel Core i7-9700F 3.0GHz 8-core CPU and 16GB DDR4 memory for all experiments. Keras \citep{chollet2015} is used on a Tensorflow 2.0 backend with Python 3.7 to train the SUHNPF networks and evaluate the MTL solvers. For optimization, we use AdaMax \citep{kingma2014adam} with parameters (\textit{lr}=$0.001$).

{\bf SUHNPF Setup}. Each training step runs for $2$ epochs, with $50$ steps per epoch. Thus, if the network takes $I$ iterations to converge, then the effective epochs taken by the network is $2I$. For computing the gradient of the Fritz-John matrix \wrt the input variables $x$, we use Tensorflow's {\tt GradientTape}\footnote{\scriptsize\url{https://www.tensorflow.org/api_docs/python/tf/GradientTape}}, which implicitly allows us to scale the computation of the gradient matrix $\nabla det$ to arbitrarily large dimensions of variable $x$. To compute the gradient update on $\mathcal{P}1$, we use a learning rate of $\eta=0.01$.

{\bf MTL Setup}. Sourcecode for LS, MOOMTL, PMTL and EPO solvers use EPO's repository\footnote{\scriptsize\url{https://github.com/dbmptr/EPOSearch}}, while EPSE\footnote{\scriptsize\url{https://github.com/mit-gfx/ContinuousParetoMTL}} and PHN\footnote{\scriptsize\url{https://github.com/AvivNavon/pareto-hypernetworks}} codes are taken from their individual repositories. 


\section{General Discussion} \label{app:discussion}


\label{sec:issues}

\textbf{Handling Non-Convex forms}: Pareto optimal solution set is a collection of saddle points \citep{van1994saddle,ehrgott2005saddle} of an MOO problem, wherein no objective can be further improved without penalizing at least one of the other objectives. This entails min-max optimization to minimize  objectives (such as loss functions) while simultaneously maximizing trade-offs between them. Although prior works \citep{sener2018multi,lin2019pareto,mahapatra2020multi} have asserted that Karush-Kuhn-Tucker (KKT) conditions \citep{boyd2004convex} in this min-max setting ensure that MTL methods find (correct) Pareto optimal solutions, it is known that KKT conditions hold true only for convex cases. \citet{gobbi2015analytical} further show that KKT-based criteria can give Pareto solutions only under fully convex setting of objectives and constraints. 

\begin{remark}
	\small The number of trade-off values that can be specified arbitrarily (by user) is dependent on the rank of matrix $L$, and is often not known a priori for a given stationary points $x^*$. Furthermore, it is not necessary that a stationary point $x^*$ exists given trade-off values $\alpha_i$s.
\end{remark}
\begin{remark}
	\small Current MTL methods with a simplex constraint on trade-off values assume that for $k$ objectives the matrix $L^TL$ is only one rank deficient or Pareto manifold is $k-1$ dimensional. These methods do not generalize to $\leq (k-2)$ dimensional Pareto manifolds or when matrix $L^TL$ is more than one rank deficient.
\end{remark}
A shown in \textbf{Appendix \ref{app:remark1}}, there can be a MOO system with $k$ objectives, where the Pareto manifold dimension is strictly less than $k-1$. Breaking up the then $k$ dimensional functional space into rays and cones is ineffective because it is trying to look for Pareto candidates in sectors, where it does not exist. More precisely, when $L\alpha=0$ and $L$ is more than one rank deficient, the trade-off values $\alpha$ have infinitely many solutions that satisfy the simplex constraints, hence for a practical solver it always becomes hard to stabilize the dual problem used in MTL methods.

\textbf{Evaluation on Benchmarks}. Because the Pareto solution is often unknown on real MOO problems, OR works have advocated that any proposed Pareto solver should first be tested on synthetic MOO with known analytic solutions. This permits controlled experimentation that vary MOO problem difficulty (\eg non-convexity in variable and function domains, presence of constraints, \etc) in order to assess the capabilities and measure the true accuracy against a known front. Ideally studies should evaluate against synthetic benchmark problems that vary in difficulty, and there is sometimes ambiguity and confusion in referring to an MOO problem as non-convex without clarifying the specific non-convex aspects. Difficulty can also vary greatly depending on whether non-convexity occurs in the objectives, constraints, or the front itself. 
\begin{remark}
	\small It is not necessary that if the objective are non-convex then the Pareto functional front is also non-convex.
\end{remark}

One typical example shown in MTL works is the double inverted gaussian benchmark, where it is stated that although the functions are pseudo-convex, the Pareto front is non-convex. However, any gradient solver is trying to find a Pareto point (\ie stationary point \wrt the objectives) of $S(x)$ in Eq. \ref{eq:linscal} for a specific alpha. As a consequence, the form of $S(x)$ (convex or non-convex) decides where the gradient ascent/descent point stabilizes. The Pareto front in the functional domain is just a post-hoc visualization of the collection of the Pareto set.

Refer to Case I, where one of the objectives is non-convex and the Pareto functional domain front is non-convex. However, in Case II, although the two functions are non-convex, the Pareto functional domain front is still convex. Although in literature the relation between the non-/convexity of the functions and the non-/convexity of the Pareto functional front has not been characterized (excluding strictly convex cases), it is an interesting direction to pursue as part of our future work.

{\bf Termination of Solvers}. An iterative solver should define termination criteria based on an error tolerance being satisfied and/or inability to further improve. It is also important that a solver reports inability to converge (achieve the termination criteria/error tolerance) within the specified maximum iterations. While both HNPF (used by SUHNPF) and EPO (used by PHN) define such error tolerance criteria for termination, inspection of source code for MOOMTL \citep{sener2018multi}, PMTL \citep{lin2019pareto}, and EPSE \citep{ma2020efficient} iterative solvers (at the time of our submission) shows support only for running a fixed number of iterations, without other termination criteria.
See the following sourcecode links to solvers for MOOMTL\footnote{\scriptsize\url{ https://github.com/dbmptr/EPOSearch/blob/master/toy\_experiments/solvers/moo\_mtl.py}}, PMTL\footnote{\scriptsize\url{ https://github.com/dbmptr/EPOSearch/blob/master/toy\_experiments/solvers/pmtl.py}}, and EPSE\footnote{\scriptsize\url{ https://github.com/mit-gfx/ContinuousParetoMTL/blob/master/pareto/optim/hvp\_solver.py}}.


\section{Additional Benchmarks} \label{app:benchmarks}

We consider two additional synthetic benchmark cases considered by \citet{navon2021learning}. We demonstrate that SUHNPF works well in these cases since the considered functions are either convex or monotone within the feasible domain for both cases.

Case A:
\begin{align*}
	&f_1(x_1,x_2) = ((x_1-1)x_2^2 + 1)/3, \, f_2(x_1,x_2) = x_2\\
	&\text{s.t.} \,\, g_1,g_2: 0 \leq x_1, x_2 \leq 1 \numberthis \label{eq:phn1}
\end{align*}

Case B:
\begin{align*}
	&f_1(x_1,x_2) = x_1, \, f_2(x_1,x_2) = 1 - (x_1/(1+9x_2))^2 \\
	&\text{s.t.} \,\, g_1,g_2: 0 \leq x_1, x_2 \leq 1 \numberthis \label{eq:phn2}
\end{align*}

Please note that although in PHN \citep{navon2021learning}, the form of $f_2=x_1$ for Eq. \ref{eq:phn1}, we believe it is a typo \wrt the original work by \cite{evtushenko2013nonuniform}, where this case was proposed, as the reported Pareto front in their work is achieved only for $f_2=x_2$. We therefore proceed with this updated form.

\begin{figure}[ht]
	\centering
	\begin{subfigure}{0.3\linewidth}
		\centering
		\includegraphics[width=\linewidth]{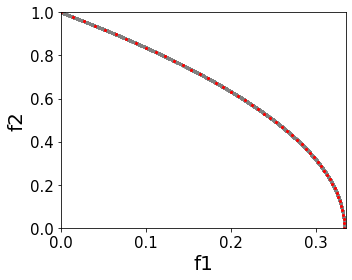}
		\caption{Case A}
	\end{subfigure}
	\begin{subfigure}{0.3\linewidth}
		\centering
		\includegraphics[width=\linewidth]{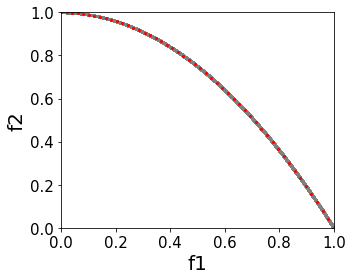}
		\caption{Case B}
	\end{subfigure}
	\caption{\small Functional Domain of cases from PHN}
	\label{fig:casesphn}
\end{figure}

\section{Loss Profiles} \label{app:loss}

\begin{figure}[ht]
	\centering
	\begin{subfigure}{0.3\linewidth}
		\centering
		\includegraphics[width=\linewidth]{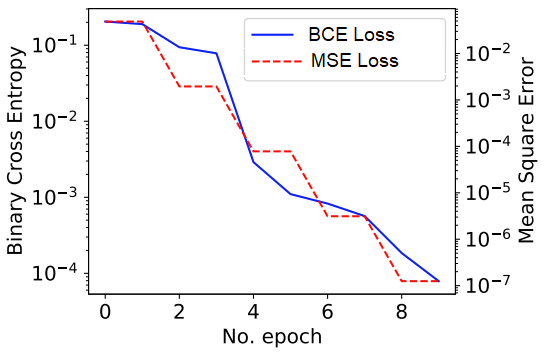}
		\caption{Case I}
	\end{subfigure}
	\begin{subfigure}{0.3\linewidth}
		\centering
		\includegraphics[width=\linewidth]{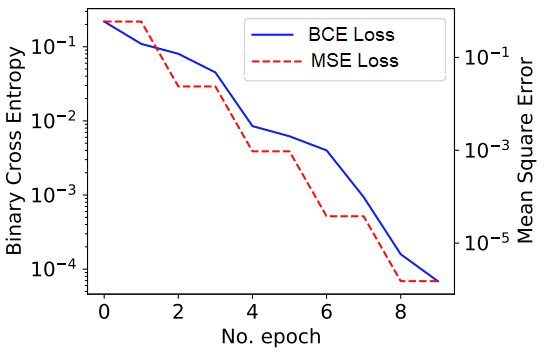}
		\caption{Case II}
	\end{subfigure}
	\begin{subfigure}{0.3\linewidth}
		\centering
		\includegraphics[width=\linewidth]{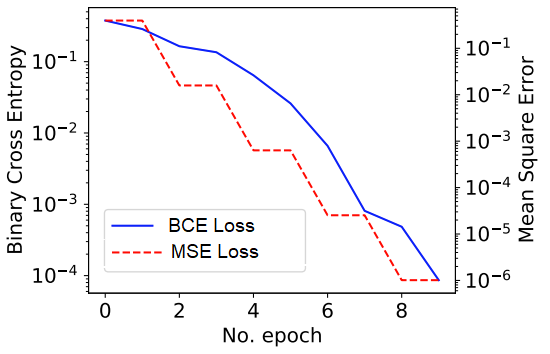}
		\caption{Case III}
	\end{subfigure}
	\caption{\small Loss profile for cases. Note that since the error threshold $\epsilon$ was set to $10^{-4}$ for the benchmark cases, the algorithm terminates once the Binary Cross Entropy (blue) loss falls below the threshold (value at epoch 10 $\leq 10^{-4}$). We also show the Mean Squared Error (dashed red) between the Pareto candidate set $\mathcal{P}1$ and the analytical solution for each iteration. Because each iteration takes two epochs, this leads to the ``staircase'' MSE shown.}
	\label{fig:loss}
\end{figure}

\textbf{Fig. \ref{fig:loss}} shows the loss profiles for the benchmark cases I - III. SUHNPF converged in $5$ iterations, with each iteration running for $2$ epochs, using error tolerance $10^{-4}$ for both the outer gradient descent loop $\epsilon_{outer}$ and inner gradient descent loop $\epsilon_{inner}$. Since the last layer of the SUHNPF network classifies points as being {\em weak} Pareto or not, the loss enforced is Binary Cross Entropy (blue line). We also report the Mean Squared Error (MSE, dashed red line) between the current iterate of point set $\mathcal{P}1$ and the true analytical solution manifold. Alg. \ref{alg:fjc} updates the Pareto candidate set in the outer descent loop. Since the inner descent loop that measures the training loss itself $\mathcal{P}1$ has ran twice for $2$ epochs, MSE is be measured only once per iteration. This results in the staircase nature of the MSE loss.

\end{document}